\documentclass[sn-mathphys,table]{sn-jnl}

\usepackage[table]{xcolor}
\usepackage[figuresright]{rotating}

\usepackage{listings}
\usepackage{color}

\definecolor{dkgreen}{rgb}{0,0.6,0}
\definecolor{gray}{rgb}{0.5,0.5,0.5}
\definecolor{mauve}{rgb}{0.58,0,0.82}

\usepackage[utf8]{inputenc}
\usepackage[T1]{fontenc}
\usepackage{placeins}
\usepackage{array}
\usepackage{booktabs}
\usepackage{tabulary}
\usepackage{amsmath, amssymb}
\usepackage{graphicx}
\usepackage{caption}
\usepackage{subcaption}
\usepackage{rotating}
\usepackage{url}
\usepackage{comment}
\usepackage{hyperref}
\definecolor{darkblue}{rgb}{0, 0, 0.5}
\hypersetup{colorlinks=true,citecolor=darkblue, linkcolor=darkblue, urlcolor=darkblue}

\jyear{2022}%

\theoremstyle{thmstyleone}%
\theoremstyle{thmstyletwo}%

\theoremstyle{thmstylethree}%

\newcommand*{\img}[1]{%
    \raisebox{-.8\baselineskip}{%
        \includegraphics[
        height=3\baselineskip,
        width=10\baselineskip,
        keepaspectratio,
        ]{#1}%
    }%
}

\begin{document}

\title[Linguistic resources]{Regionalized models for Spanish language variations based on Twitter}


\author[1,2]{\fnm{Eric S.} \sur{Tellez}}\email{eric.tellez@infotec.mx}

\author*[3]{\fnm{Daniela} \sur{Moctezuma}}\email{dmoctezuma@centrogeo.edu.mx}
\equalcont{These authors contributed equally to this work.}

\author[1,2,4]{\fnm{Sabino} \sur{Miranda}}\email{sabino.miranda@infotec.mx}
\equalcont{These authors contributed equally to this work.}

\author[1,2]{\fnm{Mario} \sur{Graff}}\email{mario.graff@infotec.mx}
\equalcont{These authors contributed equally to this work.}

\author[1,3]{\fnm{Guillermo} \sur{Ruiz}}\email{lgruiz@centrogeo.edu.mx}
\equalcont{These authors contributed equally to this work.}

\affil[1]{\orgdiv{Conacyt}, \orgname{Consejo Nacional de Ciencia y Tecnología.}, \orgaddress{\street{Av. Insurgentes Sur 1582, Col. Crédito Constructor.}, \postcode{03940}, \state{CDMX}, \country{Mexico}}}

\affil[2]{\orgdiv{INFOTEC}, \orgname{Centro de Investigación e Innovación en Tecnologías de la Información y Comunicación}, \orgaddress{\street{Circuito Tecnopolo Norte, No.112 Col. Tecnopolo Pocitos II}, \city{Aguascalientes}, \postcode{20326}, \state{Aguascalientes}, \country{Mexico}}}

\affil[3]{\orgdiv{CentroGEO}, \orgname{Centro de Investigación en Ciencias de Información Geoespacial.}, \orgaddress{\street{ Circuito Tecnopolo Norte, No.107 Col. Tecnopolo Pocitos II}, \city{Aguascalientes}, \postcode{20313}, \state{Aguascalientes}, \country{Mexico}}}

\affil[4]{\orgdiv{UPIITA-IPN}, \orgname{Instituto Politécnico Nacional}, \orgaddress{\street{Av. Instituto Politécnico Nacional 2580 Col. Barrio la Laguna Ticomán}, \city{Gustavo A. Madero}, \postcode{07360}, \state{Mexico City}, \country{Mexico}}}



\abstract{Spanish is one of the most spoken languages in the world. Its proliferation comes with variations in written and spoken communication among different regions. Understanding language variations can help improve model performances on regional tasks, such as those involving figurative language and local context information.
This manuscript presents and describes a set of regionalized resources for the Spanish language built on four-year Twitter public messages geotagged in 26 Spanish-speaking countries. We introduce word embeddings based on FastText, language models based on BERT, and per-region sample corpora. We also provide a broad comparison among regions covering lexical and semantical similarities and examples of using regional resources on message classification tasks.
}




\maketitle

\section{Introduction}
Communication is, at its core, an understanding task. Understanding a message implies that peers know the vocabulary and structure; i.e., the receiver obtains what the sender intended to say. 
Language is a determinant factor in any communication. Even people who speak the same language can find difficulties communicating information due to slight language variations due to regional variations, language evolution, cultural influences, and informality, to name a few.

A dialect is a language variation that diverges from its origin due to several circumstances. Dialects can differ regarding their vocabulary, grammar, or even semantics. The same sentence can be semantically different among dialects. In contrast, people of different dialects may need help understanding sentences with the same meaning. This effect is notoriously complex for figurative language since it contains cultural and ideological references. Studying these dialects can help us understand the cultural aspects of each population and the closeness between them.
In this sense, dialectometry studies the regional distribution of dialects. Similarly, dialectometric analysis has as its objective, through a computational approach, to analyze this distribution and provide a quantitative linguistic distance between each dialect or region of it~\cite{Donoso2017DialectometricAO}.
Hence, the research in dialectology tries to understand language differences, innovations, or variations not only in space but also in time through several phenomena. 
Natural Language Processing (NLP) tools can analyze written communications automatically. However, the support for regional languages is still in its early stages in NLP, particularly for languages different from English.

Another important aspect to discuss is about if dialects are directly related to space or geographical boundaries. As Penny~\cite{penny2000variation} discuss that language people who live in different territory do not speak the same way because the neighboring locality plays a key role. Also, the social and historical language variations are discussed in ~\cite{penny2000variation}. In some way, with the analysis done we hope to contribute a little to this exciting discussion. A more linguistic comparison from Spanish variants in America could be found in \cite{cotton1988spanish}.

On the other side, social media is a crucial component of our lives; Facebook, Twitter, Instagram, and Youtube are among the most used social networks that allow interaction among users in written form and other media. In particular, Twitter is a micro-blogging platform where most messages are intentionally publicly available, and developers and researchers can access these messages through an API (Application Programming Interface). Twitter's messages are known as {\em tweets}. Each tweet contains text and additional metadata like the user that wrote it and the geographic location where it was published. 
In the case of social media, where the source of the messages is informal, the errors are another source of variability in the language. This kind of messaging may impose extra difficulties than in formal written documents.

Nonetheless, Twitter messages' quality and socio-demographic representativeness have been continuously questioned~\cite{CR2013}. Some authors have shown that despite the over-representation of some social groups, social media usage can still be of enormous usefulness and quality~\cite{huang2016understanding}. Language and geographical information are crucial to understanding the geographies of this online data and how some information related to economic, social, political, and environmental trends could be used~\cite{graham2014world}. 

It is easy to observe the relevance of having quality and realistic data for language variations analysis or model generation with this background. Sometimes, it is difficult to have a regional-specific corpus. A large corpus is necessary to learn language models and achieve confident analysis; more data often imply better models.
With this kind of resources acquired, for instance, from social media platforms, many potential research and applications have been made in many research areas, such as health~\cite{paul2011you}, environmental issues~\cite{mooney2009evaluating}, emotion~\cite{suhasini2020emotion}, mental health~\cite{finfgeld2015twitter}, gender~\cite{vashisth2020gender}, and misogyny~\cite{frenda2019online}, among others.

There are several examples of linguistic resources attending specific tasks or applications, such as the study of Down syndrome in ~\cite{escudero2022prautocal}. Also, on~\cite{kennedy2022introducing}, a corpus is proposed for hate-based rhetoric, or hate recognition. On~\cite{gruszczynski2022electronic}, the authors provide a corpus of up to 13.5 million tokens of Polish texts between 1601 and 1772. As can be seen, a variety of corpus has been generated by the community to deal with general or more specific NLP tasks.

On the other hand, the Spanish language variations are also studied in the NLP research community. 
For instance, in~\cite{gonccalves2014crowdsourcing}, authors present a crowdsourcing language diatopic variation using Twitter data with geolocation, employing tweets messages in Spanish for more than two years over the globe. The analysis was made with a set of pre-established words and counting each and its variations worldwide. In this sense, to know what regions are close to each other, the authors used the {\em k-means} clustering algorithm over these words frequencies and PCA (Principal Component Analysis) to reduce data into two-dimensional space for visualization. The clustering approach identifies large macro-regions sharing language characteristics. 

Unfortunately, for the Spanish language, there are few works, contrary to some language variations from Europe, e.g., English dialects~\cite{hovy2020visualizing}, 
French dialects~\cite{lamontagne2022phonological} and Arabic~\cite{alshutayri2017exploring}, to name some. 

One of the possible assumptions using Twitter is that the behaviors of English users generalize to other language users. In~\cite{hong2011language} is presented a study using 62 million tweets over more than 100 different languages over four weeks. Applying an automatic language detection algorithm, they found that most of the data were in English (51\%), and 39\% was for other languages such as Japanese,  Portuguese,  Indonesian,  and  Spanish.

The geographical region of a language helps to know how this language is used in a particular society. For instance, Spanish is a largely used language; nevertheless, it is used differently according to the country or even a more specific geographical location. Hence, the language could analyze at the regional level~\cite{huang2016understanding, rodriguez2018dialectones}.
In~\cite{rodriguez2018dialectones}, the authors study Spanish language variations in Colombia. The analysis used unigram features, and the authors stated that it was challenging to compare Spanish variations against regions identified by other authors using classical dialectometry. Hence, in conclusion, the authors said that automatic detection of {\em dialectones} is an adequate alternative to classical methods in dialectometry for automated language applications.
A more current effort to deal with the Spanish Language on Twitter was presented by \cite{huertas2022bertuit}, which provides a powerful tool to take advantage of transformers generated with the native language. The authors test their solution with several classical NLP tasks.

The number of articles related to other languages is also reduced. In~\cite{alshutayri2017exploring}, the Arabic dialects were classified using the WEKA tool\footnote{https://www.cs.waikato.ac.nz/ml/weka/} reaching an accuracy of 79\% on their classification results. The used dataset contained 210,915K tweets from some Arabic dialects and classified them considering their geographic location.

Mocanu et al.~\cite{mocanu2013twitter} survey the linguistic landscape in the world using Twitter. This landscape includes linguistic homogeneity or variations over countries that consider the touristic seasons. The method employed to identify language is the Chromium Compact Language Detector by Google; the authors also used the location of the devices reported by tweets. As a result, it was possible to observe distributions of language on Twitter over several countries by month of the year and where touristic flow is evident.

Also, emoticons or emojis are effective communication symbols. Their usage has been studied in the literature. For instance, in~\cite{park2013emoticon}, authors analyzed the semantic, cultural, and social aspects of their use on Twitter.
Kejriwal et al.~\cite{kejriwal2021empirical} studied the use of emojis in terms of linguistic use and countries. The authors collected tweets from 30 different languages and countries, and the authors found that emojis usage strongly correlates between language and country level, which means that emojis are used according to language and region. Another example of studying Emojis is presented in ~\cite{li2019empirical}.

Other efforts have been made to exploit the regionalized models for a specific language variation; for instance, in~\cite{jimenez2018automatic} where methods to identify regional words and provide their meaning is studied.

Our contribution is a set of regionalized resources for different variations of the Spanish language. We created and characterized regional vocabularies and regional semantic representations of them (i.e., word embeddings). Also, we learn and test language models based on BERT. We built these resources from an extensive collection of public tweets from 2016 to 2019, written in 26 countries with a large basis of Spanish-speaking people. Regarding messages, we provide a sample of Twitter message identifiers divided by region such that researchers can retrieve them easily. Finally, we show some usage examples of our resources.

The rest of the manuscript is organized as follows. 
Section \ref{sec:dataset} describes our Twitter Spanish Corpora (TSC), used to generate our regional resources.
Section~\ref{sec:lexical-analysis} compare lexical traits among the corpora. Section~\ref{sec:semantic-analysis} is dedicated to presenting our semantic resources and their affinity analysis that includes visualizations and experimental evidence that support the use of regional word embedding models on regional tasks. Our resources based on language models are presented and compared in Section~\ref{sec/language-models}. Finally, Section \ref{sec:conclusions} summarizes and discusses the implications of the TSC.

\section{Twitter corpora of the Spanish language}
\label{sec:dataset}
With 489 million native speakers in 2020\footnote{https://blogs.cervantes.es/londres/2020/10/15/spanish-a-language-spoken-by-585-million-people-and-489-million-of-them-native}, Spanish is one of the languages with a higher native speaking basis, just ranked behind Chinese Mandarin in terms of the number of native speakers.
Twenty-one countries have the Spanish language as the official language (by law or \textit{de facto})\footnote{https://en.wikipedia.org/wiki/List\_of\_countries\_where\_Spanish\_is\_an\_official\_language}.
Our corpora selected these regions, see Table~\ref{tab/country-properties}; and we also considered five additional regions (US, CA, GB, FR, and BR) with well-known migration, business, and tourism activities of Spanish speakers.
The number of Twitter users varies with each country; since each country has different social, political, security, health, and economic conditions, we will avoid generalizations.

As mentioned, we collected publicly published tweets between 2016 and 2019 using the Twitter stream API. Also, we limited our collection to geotagged messages marked by Twitter as written in Spanish.
We decided to let out the corpus messages from the year 2020 and posteriors to avoid disturbances in social media regarding the COVID-19 pandemic. Please recall our objective is to build resources based on the language itself and not analyzing the pandemic event. Twitter stream API allows tweet retrieval in two ways. The first consists of using a language marker ({\em  lang=es}, for Spanish) and a list of tracking words linked to the specified language. In this case, we can use Spanish {\em stopwords}\footnote{Words that are so common in a language, such as articles, prepositions, interjections, and auxiliary verbs, among other typical words.} to maximize the download process. The second strategy consists in using a language marker ({\em lang=es}, for Spanish) and geographical coordinates, these kinds of tweets are named geotagged Tweets. We specify worldwide coordinates to get tweets from everywhere. We use only geotagged tweets from the last strategy. These geotagged tweets have information such as country code corresponding to the country where the tweet was published, among other metadata. We rely on the information provided by Twitter about the country associated with each tweet.

To ensure a minimum amount of information in each tweet, we discard those tweets with less than five tokens, i.e., words, emojis, or punctuation symbols, following the strategy of \cite{mikolov2013distributed} for analyzing and learning from very large collections. We also removed all retweets to avoid duplication of messages and reduce foreign messages commented on by Spanish speakers. After this filtering procedure, we retain close to $800$ million messages.

\begin{table}[!ht]
\caption{Datasets' statistics after filtering by retweets and ensuring at least five words per tweet. We show the origin country, the country code in ISO 3166-1 alpha-2 format reported by the Twitter API, the number of tweets, and the number of different users in the collected period.}
\label{tab/country-properties}
\vspace{0.07cm}
\centering
\resizebox{\textwidth}{!}{
\begin{tabulary}{1.2\textwidth}{R>{\tt}RRR RRR}
\toprule
\bf country &
\bf code &
\bf $\alpha$ & 
\bf $\beta$ & 
\bf number of users &
\bf number of tweets &
\bf number of tokens
\\ \midrule
Argentina          & AR & 0.7563 & 1.8594 & 1,376K & 234.22M & 2,887.92M  \\
Bolivia            & BO & 0.7509 & 1.8913 & 36K    &  1.15M  &    20.99M  \\
Chile              & CL & 0.7555 & 1.8874 & 415K   & 45.29M  &   719.24M  \\
Colombia           & CO & 0.7562 & 1.8993 & 701K   & 61.54M  &   918.51M  \\
Costa Rica         & CR & 0.7447 & 1.8595 & 79K    &  7.51M  &    101.67M \\
Cuba               & CU & 0.7640 & 1.8677 & 32K    &  0.37M  &     6.30M  \\
Dominican Republic & DO & 0.7544 & 1.8832 & 112K   &  7.65M  &    122.06M \\
Ecuador            & EC & 0.7538 & 1.8968 & 207K   & 13.76M  &    226.03M \\
El Salvador        & SV & 0.7494 & 1.9066 & 49K    & 2.71M   &    44.46M  \\
Equatorial Guinea  & GQ & -      & -      & 1K     & 8.93K   &     0.14M  \\
Guatemala          & GT & 0.7498 & 1.9175 & 74K    & 5.22M   &    75.79M  \\
Honduras           & HN & 0.7486 & 1.8941 & 35K    & 2.14M   &    31.26M  \\
Mexico             & MX & 0.7557 & 1.8895 & 1,517K & 115.53M & 1,635.69M  \\
Nicaragua          & NI & 0.7445 & 1.8535 & 35K    & 3.34M   &    42.47M  \\
Panama             & PA & 0.7559 & 1.8952 & 83K    & 6.62M   &    108.74M \\
Paraguay           & PY & 0.7511 & 1.8815 & 106K   & 10.28M  &   141.75M  \\
Peru               & PE & 0.7583 & 1.8966 & 271K   & 15.38M  &   241.60M  \\
Puerto Rico        & PR & 0.7498 & 1.8433 & 18K    & 0.58M   &     7.64M  \\
Spain              & ES & 0.7648 & 1.9036 & 1,278K & 121.42M & 1,908.07M  \\
Uruguay            & UY & 0.7516 & 1.8346 & 157K   & 30.83M  &   351.81M  \\
Venezuela          & VE & 0.7614 & 1.8959 & 421K   & 35.48M  &   556.12M  \\
\midrule
Brazil                   & BR & 0.7681 & 1.9389 & 1,604K & 27.20M &  142.22M  \\
Canada                   & CA & 0.7652 & 1.9331 & 149K   & 1.55M  &  21.58M  \\
France                   & FR & 0.9372 & 1.9324 & 292K   & 2.43M  &  27.73M  \\
Great Britain            & GB & 0.7687 & 1.9129 & 380K   & 2.68M  &  34.62M  \\
United States of America & US & 0.7666 & 1.8929 & 2,652K &40.83M  & 501.86M \\ 
\midrule
Total                    &    &        &        & 12M        &  795.74M &  10,876.25M \\
\bottomrule
\end{tabulary}
}
\end{table}

Table~\ref{tab/country-properties} shows statistics about our corpora describing aspects such as country, number of users, number of tweets, and number of tokens. The table shows that Spain, the USA, Mexico, and Argentina are countries with more users. Furthermore, they are also those with more tweets in the Spanish language, but the USA falls considerably in this aspect. A similar proportion is observed in the number of tokens column. However, Argentina has the highest number of tokens, above Mexico and Spain significantly.

The table also lists the coefficients for the expressions behind Heaps' and Zipf's laws. Both laws are broadly surveyed in the literature; for instance, \cite{Gelbukh2001} and \cite[Chapter~5]{schutze2008introduction} describe them and study their implications from a general perspective.
In a nutshell, the laws describe how the vocabulary grows in text collections written in non-severe-agglutinated languages. Heaps' law $n^\alpha$ describes the sub-linear growth of the vocabulary on a growing collection of size $n$. Zipf's law represents a power-law distribution where a few terms have very high frequencies, and many words occur with a shallow frequency in the collection. The expression that describes Zipf's law is $1 / {r^\beta}$, where $r$ is the rank of the term's frequency.

\begin{figure}
\centering
\begin{subfigure}[t]{0.5\textwidth}
        {\includegraphics[trim=0.5cm 0cm 1.9cm 0.5cm, clip=true, width=\textwidth]{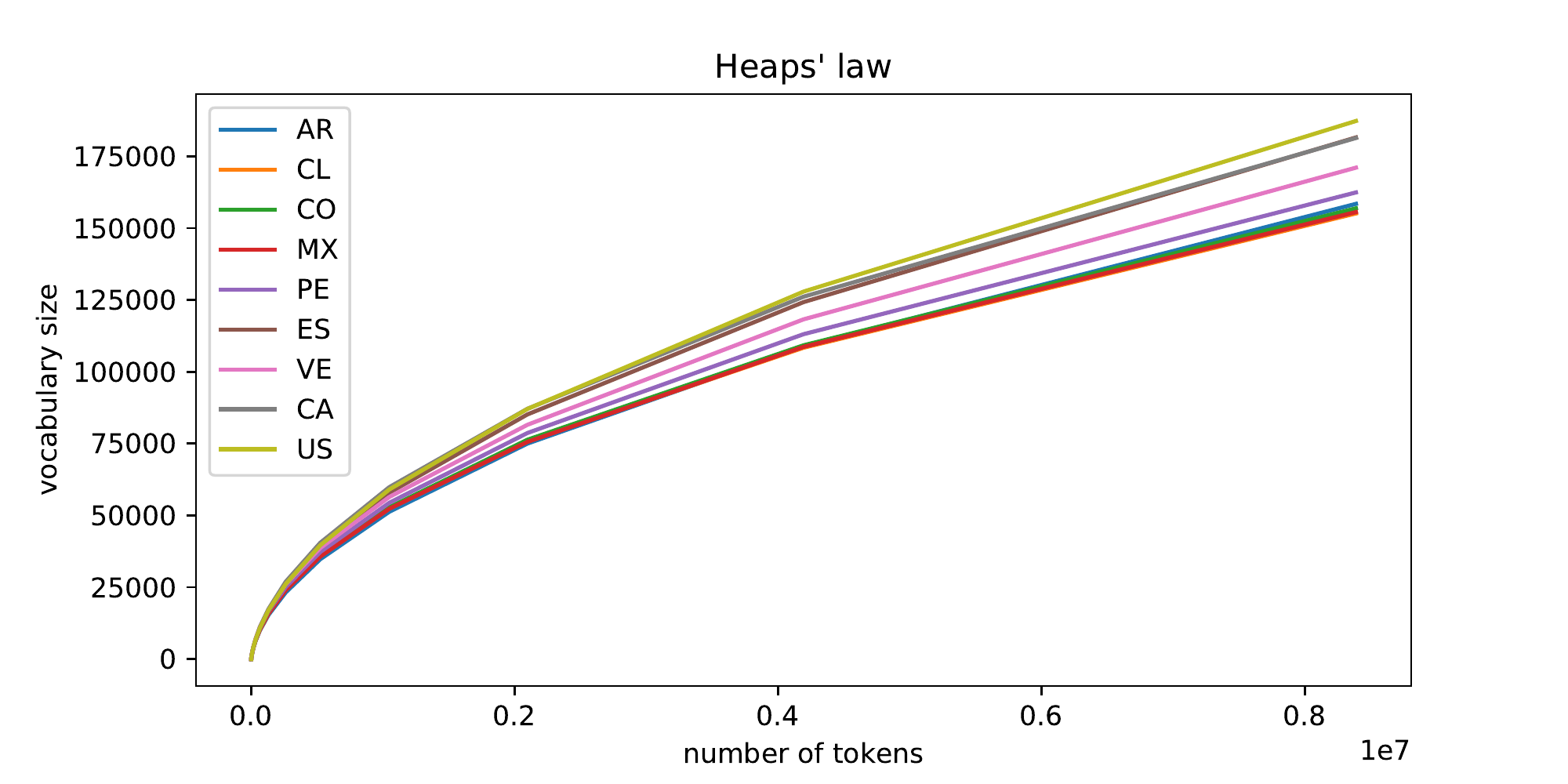}}
    \caption{Heaps law. Vocabulary size with respect to the number of tokens.}
    \label{fig:heaps}
\end{subfigure}~\begin{subfigure}[t]{0.5\textwidth}
    {\includegraphics[trim=.5cm 0cm 1.9cm 0.5cm, clip=true, width=\textwidth]{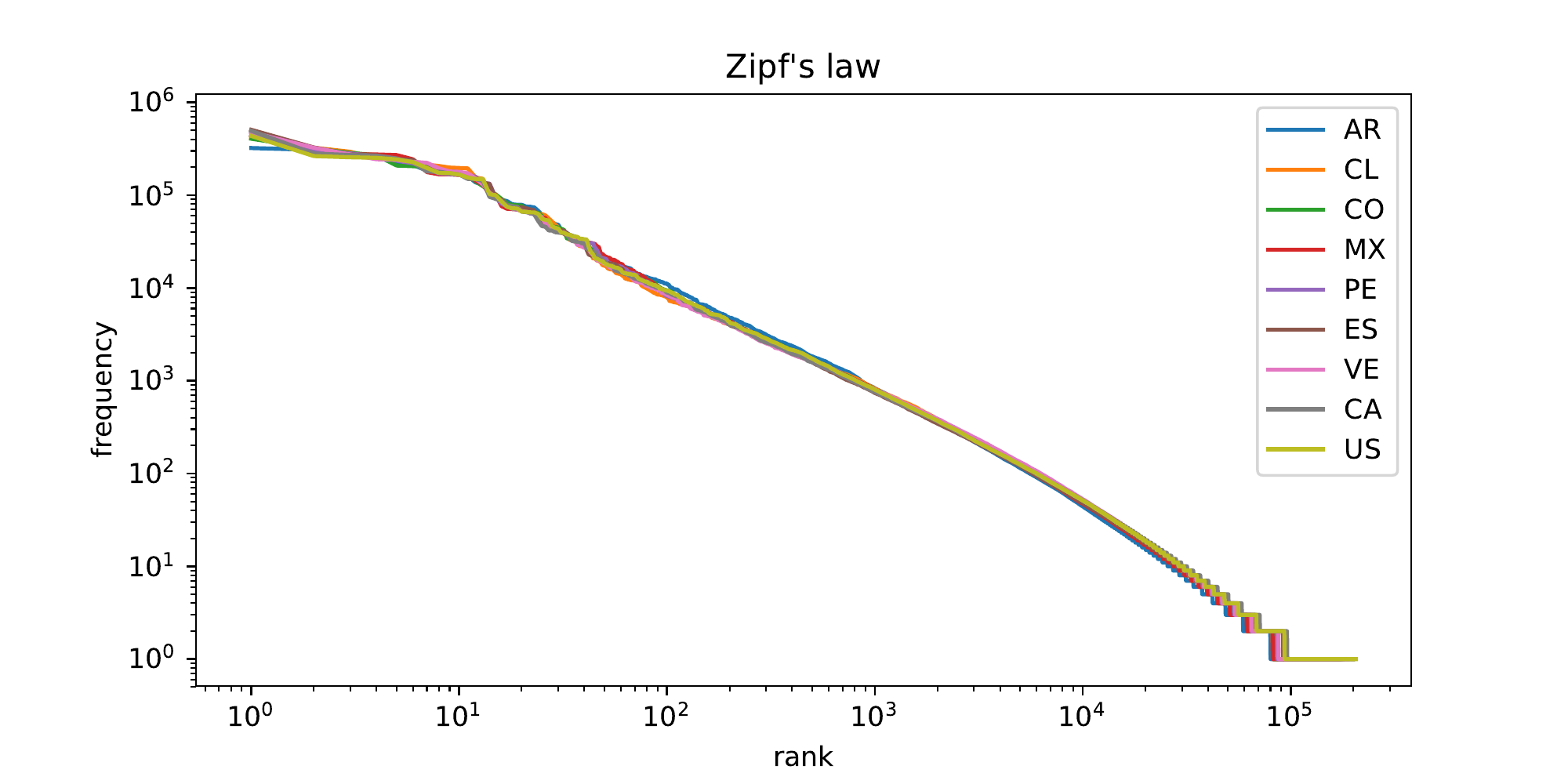}}
    \caption{Zipf's law. Frequency of tokens.}
    \label{fig:zipf}
\end{subfigure}
\caption{The vocabulary growth and distribution of frequencies of $10^7$ tokens over a sample of our Twitter's Spanish language corpora.}
\label{fig:heaps-zipf}
\end{figure}

Figure~\ref{fig:heaps} illustrates the Heaps' law in a small sample of regions of interest. One can observe its predicted sub-linearity and that Mexico has the lowest growth in its vocabulary size according to the number of tokens. On the contrary, the US corpus shows faster vocabulary growth, possibly explained due to the mix of languages in many messages.

Figure~\ref{fig:zipf} shows Zipf's law under a log-log scale and its quasi-linear shape. We can see slight differences among curves, more noticeable on both the left and right parts of the plot. The left part of the curves corresponds to those terms with very high frequency, and the right side is dedicated to those terms being rare in the collection. Notice that all these curves are similar but slightly different; this is not a surprise since we analyze variations of the same idiom, i.e., the Spanish language. 

\subsection{Geographic distribution}
\begin{figure}[t]
    \begin{subfigure}[t]{0.5\textwidth}
	{\includegraphics[trim=2cm 1cm 5.95cm 1.2cm, clip=true, width=1\textwidth]{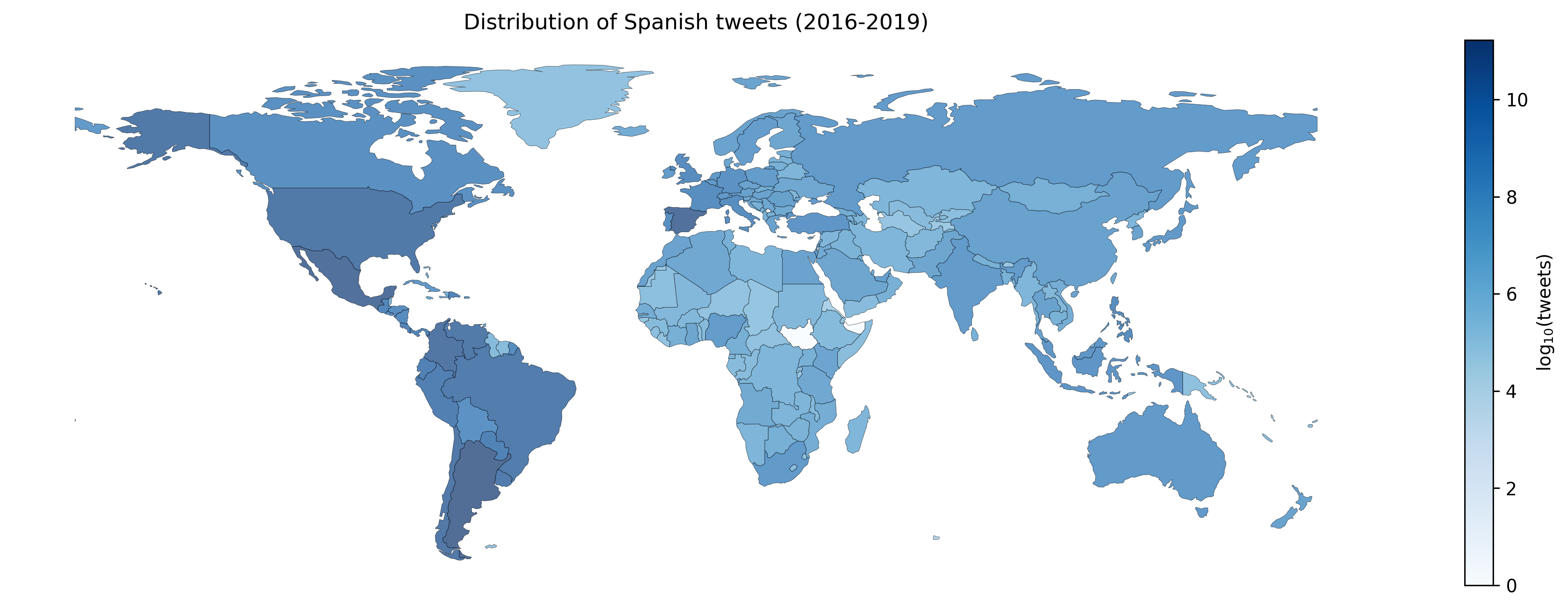}}
	\caption{Distribution of tweets tagged as Spanish by Twitter.}
	\label{fig:tweets}
	\end{subfigure}~~\begin{subfigure}[t]{0.5\textwidth}
	{\includegraphics[trim=2cm 1cm 5.95cm 1.2cm, clip=true, width=1\textwidth]{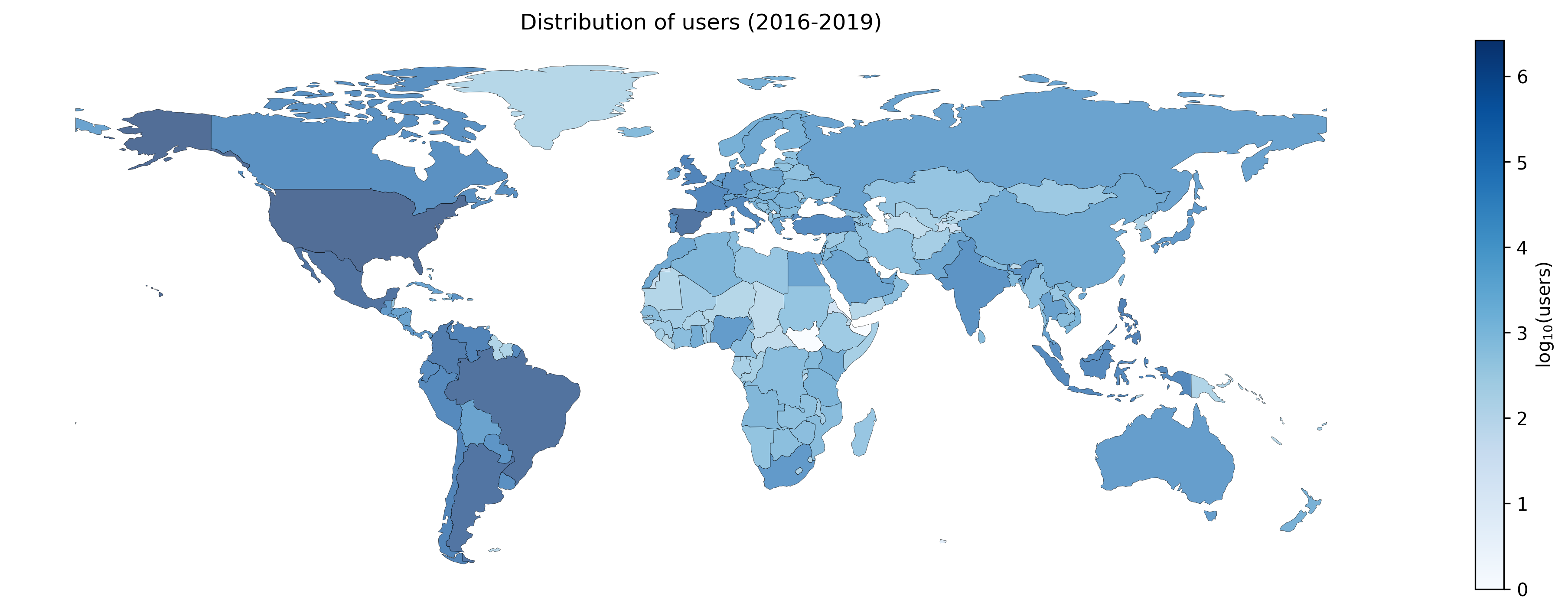}}
	\caption{Distribution of the number of tweeters with at least one tweet tagged as Spanish language.}
	\label{fig:tweeters}
	\end{subfigure}
	
	\caption{Distribution of tweets and tweeters labeled as Spanish-speaking users around the world. Colors are related to the logarithmic frequencies in data collected from 2016 to 2019 with the public Twitter API stream. Darker colors indicate a high population; the logarithmic scale implies that only significant frequency differences produce color changes.}
	\label{fig:geo-distributions}
\end{figure}

Figure~\ref{fig:tweets} illustrates the number of collected Spanish language tweets over the world.
The color intensity is on a logarithmic scale, which means that slight variations in the color imply significant changes in the number of messages; countries with the darkest blue have the highest number of tweets in Spanish. This figure shows how American countries (in the south, central, and north) and the Iberian Peninsula have, as expected, more tweets in the Spanish language than the rest of the world.

Figure~\ref{fig:tweeters} shows the distribution of tweeters (users) per country. As in the previous image, we present a logarithmic scale in the intensity of color to represent the number of users. The differences between this figure and Figure~\ref{fig:tweets} are low, as expected, and follow the same distribution. Note the high intensity of American countries.

\begin{figure}
    \centering
    \includegraphics[width=0.7\textwidth]{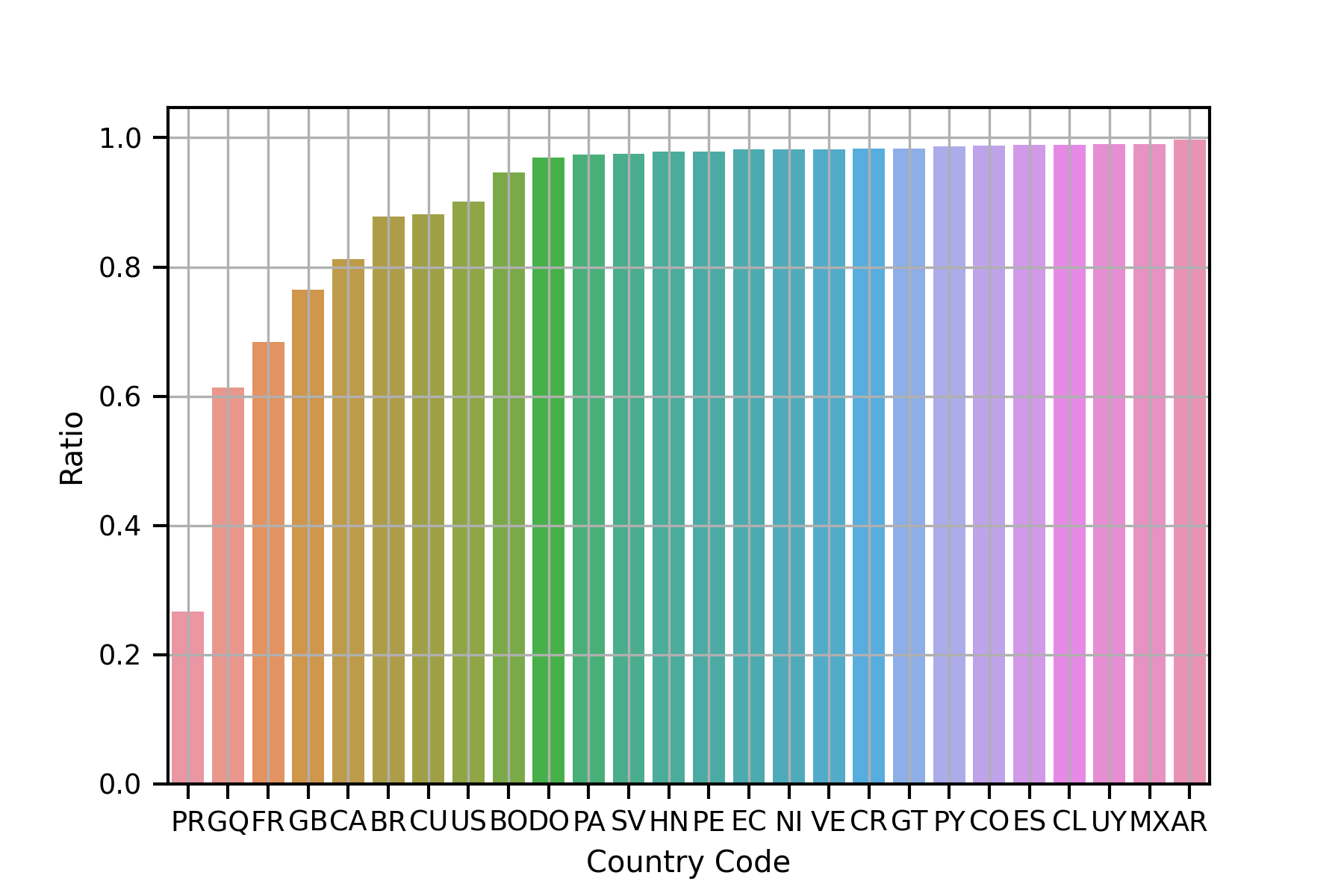}
    \caption{Ratio between the number of tweets produces by local tweeters and the total tweets in the country.}
    \label{fig:prop_tweets_in_country}
\end{figure}

Also, it is essential to know whether the tweets come from persons living in the country or travelers. It might be impossible to completely answer this by looking only at the tweets. However, an approach that can provide insight into the locality of the tweets is to measure the ratio between the tweets produced by local users and the total number of tweets produced in each country, where each user is assigned to the country where more tweets have. For example, suppose a tweeter has 100 tweets in Mexico and 10 in Spain. In that case, that user is considered Mexican, and those published in Spain (considered there a tourist) are not considered, which means only the tweets produced in Mexico are counted.  

Figure \ref{fig:prop_tweets_in_country} presents the ratio between local tweets and the total number of tweets; it can be observed that Puerto Rico (PR) has the lowest ratio, meaning that most tweets come from foreigners. The second lowest is Equatorial Guinea (GQ), where more than 60\% of its tweets are local. Then, there is a block of countries where Spanish is not the primary language. 
After the block comes Cuba, with 88\% of its tweets produced by locals. The United States of America (US) has a ratio of 90\%. From the Dominican Republic (DO) to the right have a percentage higher than 95\%. The median ratio in these countries is 98\%.
This effect could indicate most of the tweets are produced by residents.


\section{Lexical resources}
\label{sec:lexical-analysis}
This section describes and analyzes our Spanish Twitter Corpora (STC) in the lexical aspect, specifically from the vocabulary usage perspective. This analysis complements that given of the Heaps' and Zipf's laws and the information given in Table~\ref{tab/country-properties}.

\begin{figure}[!ht]
\centering
    \includegraphics[width=0.95\textwidth]{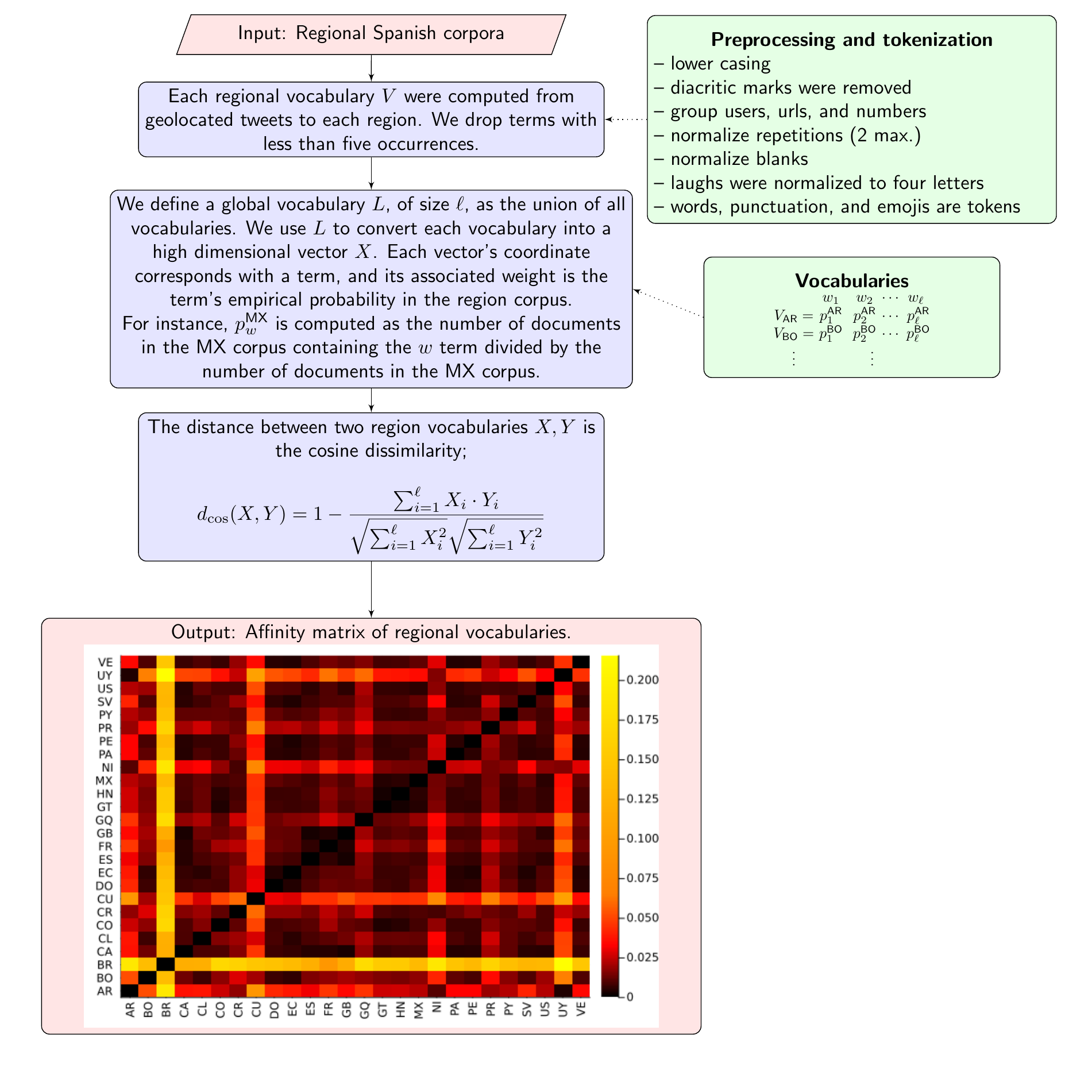}
    \caption{Affinity matrix among Spanish regions' vocabularies.}
    \label{fig:lex-affinity-matrix}
\end{figure}

Figure~\ref{fig:lex-affinity-matrix} describes the procedure applied to obtain an affinity matrix of our Spanish corpora.\footnote{Under our context, an affinity matrix is a pair-wise matrix of distances among different regions using a dissimilarity function. The $i$th row contains the distance of the $i$ region vs. all regions; it has a zero diagonal.} For this purpose, we extracted the vocabulary of each corpus, i.e., a matrix that describes the similarities among corpora. The vocabulary was computed on the entire corpus after text normalizations described in the diagram. We also removed those terms with less than five occurrences in the corpus to remove the tail of the term-frequency distribution, similarly to \cite{mikolov2013distributed,bojanowski2017enriching}. The remaining terms are used to create a vector that represents the regional corpus.

The affinity matrix is computed using the cosine distance described in the flow diagram. Note that we select the cosine distance as metric due to the reminiscences of the traditional bag of words with our vocabulary representation, see \cite{schutze2008introduction} for more information about conventional vector models for information retrieval. The heatmap represents the actual values in the matrix. This matrix is crucial for the rest of this analysis since it contains distances (dissimilarities) among all pairs of our Spanish corpora.
Values close to zero (darker colors) imply that those regions are pretty similar, and lighter ones (close to one) are those regions with higher differences in their vocabularies.
For instance, the affinity matrix can show us how Mexico (MX) is more similar to Honduras (HN), Nicaragua (NI), Peru (PE), and the USA (US). This behavior could be the geographical location of the countries, and therefore, a large migration or cultural interchange is made. On the other hand, Brazil (BR) and Equatorial Guinea (GQ) are among the most atypical countries with low similarities with the other countries. 

\begin{figure}[!ht]
\centering
    \centering
    \fcolorbox{gray}{white}{\includegraphics[width=0.6\textwidth]{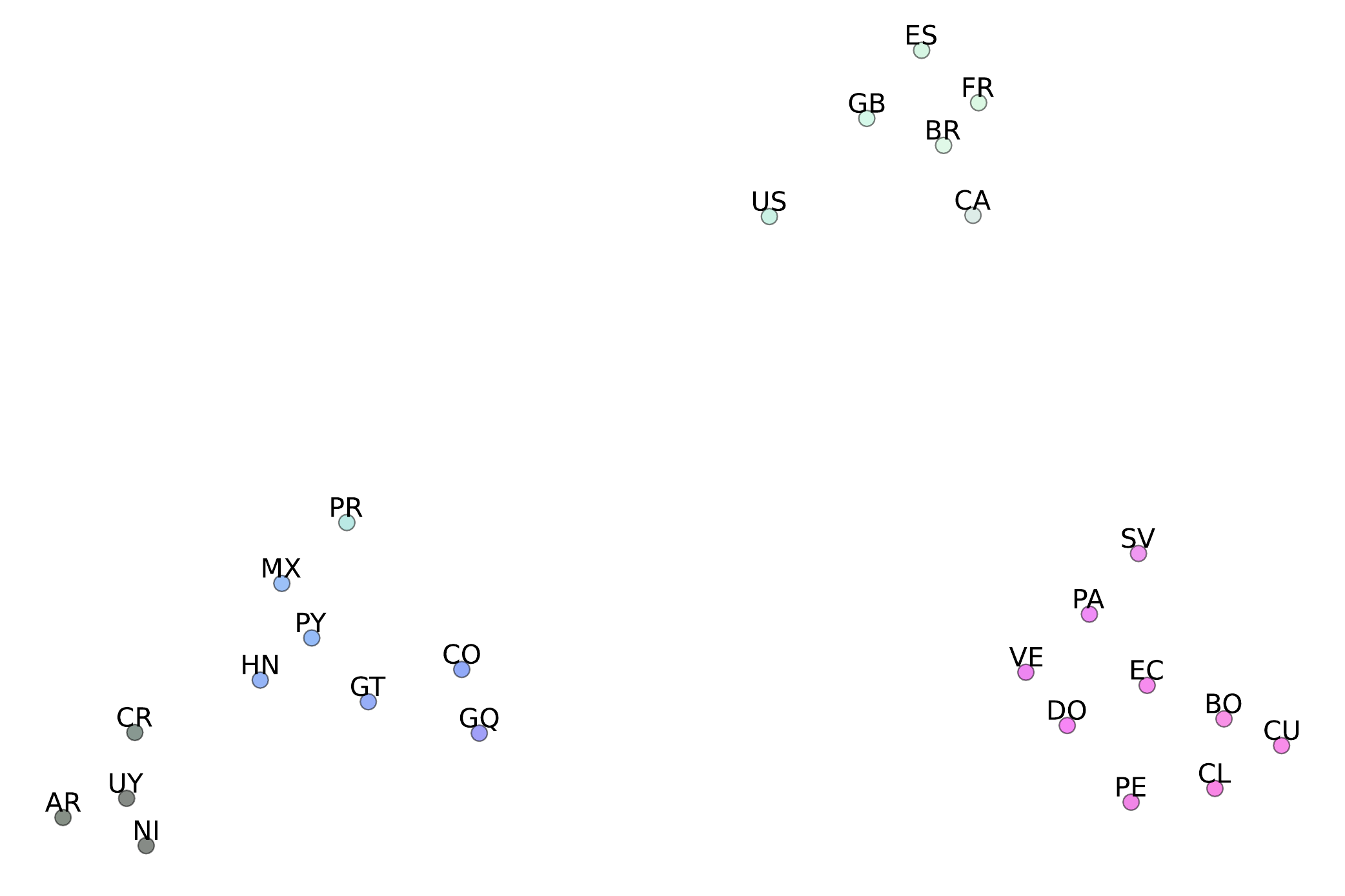}}


\caption{Spanish-language lexical similarity visualization among country's vocabularies through a two-dimensional UMAP projection using the Cosine among vocabularies. The points were colorized using a 3D UMAP projection (normalized and interpreted as RGB). Both projections use three nearest neighbors, which emphasizes local features.}
\label{fig:lex-voc-umap}
\end{figure}

Figure~\ref{fig:lex-voc-umap} illustrates the similarity between Twitter country vocabularies.
Here we rely on Uniform Manifold Approximation and Projection (UMAP) algorithm~\cite{mcinnes2020umap}, a non-linear dimension reduction technique that approximates the $k$ nearest neighbor graph structure of a dataset in the projected low dimension. We applied UMAP projections (2D for spatial projection and 3D for colorizing points) using the affinity matrix as input. Please recall that the affinity matrix represents similarities between regional vocabularies using the cosine distance as the metric. Please remember that the UMAP algorithm uses the affinity matrix to generate the low-dimensional projection. To our best knowledge, this is a novel approach to visualizing similarities among vocabularies. Still, we can find several uses of dimensional reduction techniques like tSNE~\cite{wada2018unsupervised} for visualizing word meaning similarities in a single multi-language model.
UMAP is a more recent non-linear dimensional reduction technique that typically performs faster than other non-linear alternatives like tSNE or ISOMAP with remarkable stability on projections~\cite{mcinnes2020umap}. Interested readers on alternatives are referenced to the recent literature~\cite{anowar2021dimension}.

The figure shows how close or far each Spanish variation is among the entire corpora. UMAP is parameterized by the number of nearest neighbors (k$nn$) in the affinity matrix.
The number of neighbors accepts values between $k=2$ and $n$, i.e., the number of elements in the collection.
Small values of $k$ capture local characteristics of the graph's structure, while large $k$ values capture global structures.
 
The figure shows the projection using 3$nn$;\footnote{The value $k=3$ was chosen after several tests, larger values capture more global characteristics, and $k=2$ produce many sparse clusters (local characteristics).} we can see four well-defined clusters here. 
For instance, Uruguay (UY) is very close to Argentina (AR) in three figures, and this is the case in other countries, like Mexico (MX), Colombia (CO), and the United States (US); or Venezuela (VE), and Ecuador (EQ).

\begin{figure}[!ht]
\centering
    \includegraphics[width=\textwidth]{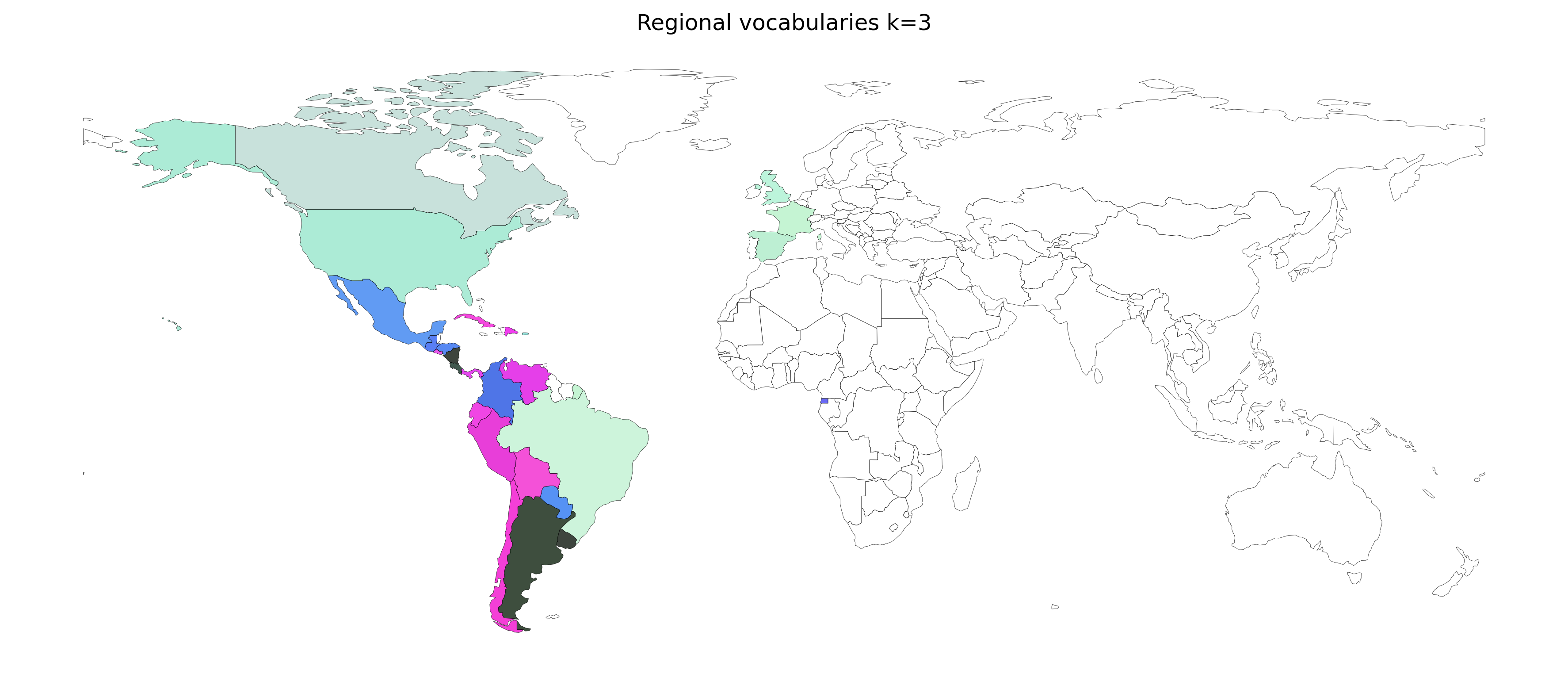}
    \caption{Regional Vocabulary in RGB representation}
    \label{fig:fig-region-lex-colors-clustering-umap-3}
\end{figure}

While some of these clusters support the idea that geographical similarities imply language similarities, there are notorious exceptions. Figure~\ref{fig:fig-region-lex-colors-clustering-umap-3} shows a colorized map using the same colors encoding of Figure~\ref{fig:lex-voc-umap}. While it is possible to observe similarities and divisions among North America, Central and South America, and European countries, there are essential differences. For instance, Colombia (CO), a South American country, has more similarities to Central American language variants.

Regarding our lexical features, Cuba and the Dominican Republic are close to Venezuela, Bolivia, and Ecuador. 
It is interesting to recall that these similarities are present in Twitter and may vary from other data sources. Still, it could be helpful to take advantage of this knowledge.

Our collections are small for some countries, e.g., GQ, PR, and CU, which can introduce some possible issues in our analysis. For instance, it is possible to declare similarities that are non-meaningful. Please recall that the UMAP projection uses the $k$ nearest neighbor graph (that takes the affinity matrix of Fig.~\ref{fig:lex-affinity-matrix} as input). Even when other regions do not select these regions as neighbors (please recall we used $k=3$), these regions will have direct neighbors that can have enough data. For instance, Fig.~\ref{fig:lex-affinity-matrix} shows Cuba with a light row column, which means that most regions are seen relatively far, but it connects Bolivia. Bolivia has more strong connections that positioned Cuba on the map. Another effect occurs with Brazil that even when it has a large corpus, it contains a lot of messages mixing Spanish and Portuguese. It is pretty different from most regions but visible closer to FR and GB, and therefore, BR will be placed near them on the 2D and 3D projections, see Fig.~\ref{fig:lex-voc-umap}.


\begin{figure}[!ht]
    \centering
    {\includegraphics[trim=6cm 1cm 3.5cm 0.5cm, clip=true, width=0.85\textwidth]{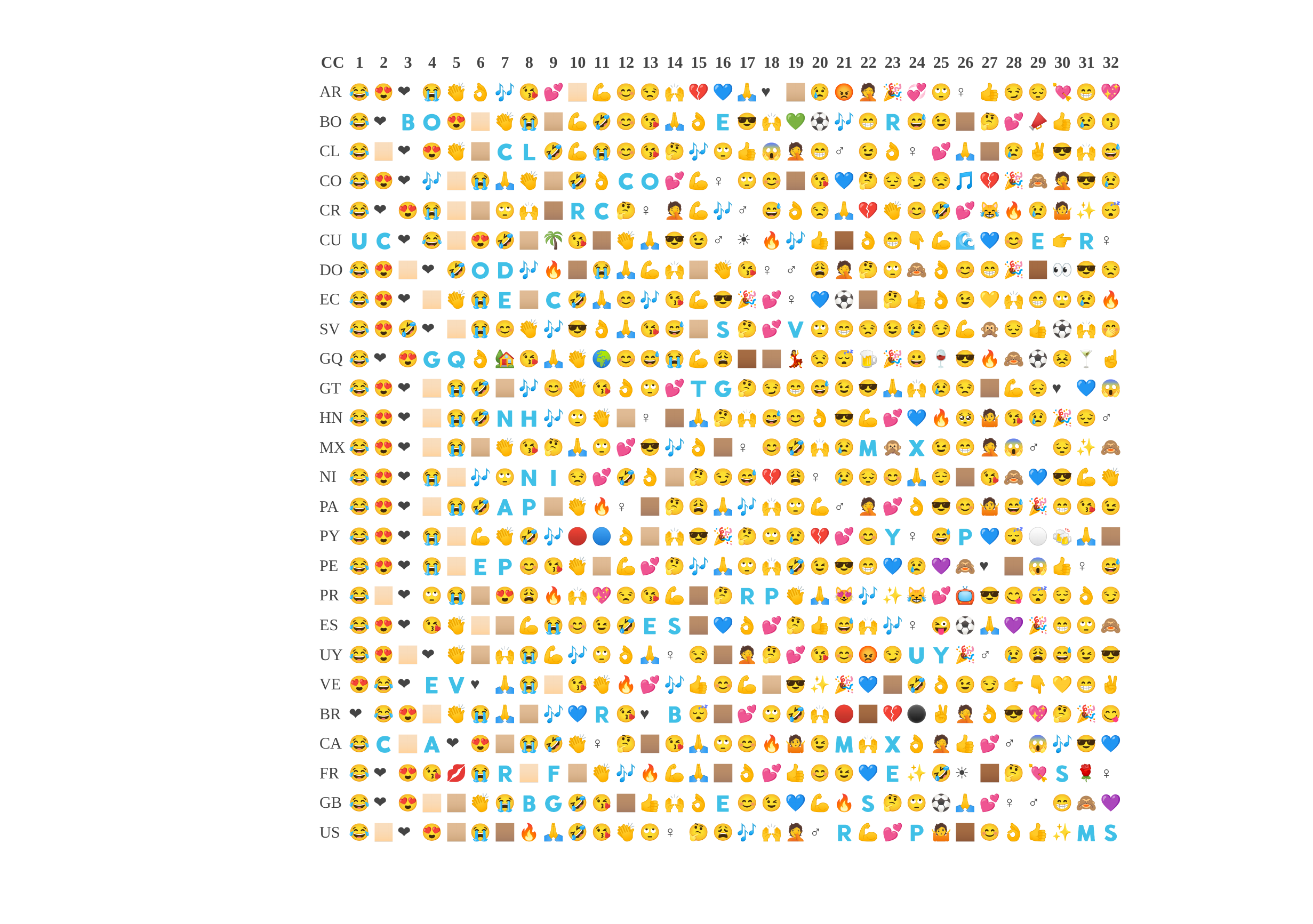}}\caption{Most popular emojis per Spanish-speaking country.}
    \label{fig:emojis}
\end{figure}

In addition, emojis are graphical symbols expressing an emotion or popular concepts. Hence, they are a lexical resource that can also imply an emotional charge. 
Emojis compensate for the lack of facial expressions and other expressive ways of face-to-face conversations. Therefore, emojis are popular on social networks like Twitter since they are concise and friendly ways to communicate~\cite{dresner2010functions}. 
The use of emojis is also dependent on the region, as illustrated in 
Figure~\ref{fig:emojis}. The figure shows the 32 most used emojis in each country; skin tone markers were separated from composed emojis and counted in an aggregated way. Note that the most popular emojis have consensus in almost all regions. In top rank, we found the {\em laughing face}, the {\em in love face}, and the {\em heart} (love). Another symbol that deserves attention is the color-skin mask, which marks emojis with a skin hue.
Regarding frequencies, lighter color-skin marks are more popular than darker ones; this information could have different meanings. For example, users identified as white people, or perhaps it is tricky to select the proper one with Twitter clients. The real reason behind this finding is beyond the scope of this manuscript but deserves attention.
\FloatBarrier

\section{Semantic analysis and regional word embeddings}
\label{sec:semantic-analysis}

This section discusses the creation of regional semantic representations (word embeddings) for our Spanish language corpora and also analyze similarities between regions using visualization techniques.
Word embeddings are vector representations of a vocabulary that capture the semantics of words learning how words are used in a large text corpus. Algorithms learns a high dimensional vector for each token using a distributional hypothesis: words used in similar contexts have similar semantics. Therefore, if two vectors are close, both are semantically related; the contrary also becomes true, as two distant vectors are different semantically.
In summary, embeddings are a popular and effective way to capture semantics from a corpus~\cite{Yang2018}. There exist several techniques to learn word embeddings, for instance, Word2Vec~\cite{Mikolov2013}, FastText~\cite{joulin2017bag,bojanowski2017enriching}, and Glove~\cite{glove}. Our resources are FastText models, to support out-of-vocabulary words, which are common in social network data.
FastText is both a word representation generator and a text classification tool. It is an open-source library well-known for its broad language coverage.\footnote{\url{https://fasttext.cc/}} For instance, Grave et al.~\cite{grave2018learning} trained word embeddings for 157 languages using Wikipedia (800 million tokens) and Common Crawl (70 billion tokens); these models include support for the Spanish language. Nonetheless, there is a lack of country-level language support to our knowledge. Our resources are the first broad effort on this matter, making it possible to take advantage of regionalisms and Spanish dialects.

We use our Twitter corpora, divided by country, and apply our preprocessing step, described in Section~\ref{sec:lexical-analysis}, as input of the FastText algorithm. Before, we filtered out several messages: we removed messages with URLs, messages with less than seven tokens, and those produced by applications that use a template to write the tweet (e.g., Foursquare). We also removed retweeted messages. These decisions emphasize that messages contain useful textual information and do not reference external data. The rationale of retweets and external data removal is that they may be of a different user, and we cannot be sure that the linked resource is located in the same region as the original message. These filters reduced our corpora in half (close to 400 million messages).

Regarding the minimum size of seven tokens, the idea is to preserve some context for each token in the message; please recall that word embeddings learn the distributional semantics of each word using its surrounding words. At the same time, we are unaware of a proper study about the phrase's minimum length required to learn word embeddings. However, based on the FastText implementation that uses sliding windows of size five by default as context, preserving messages with at least seven tokens is a tradeoff between maintaining a large dataset and filtering out very short messages.

As commented, we created 26 word-embedding models, one per country, and learned 300 dimension vectors, which is almost a standard for pre-trained embeddings. We used the default values for the rest of the hyper-parameters of FastText. In addition, for comparing purposes, we use the entire corpora as a single corpus to create global word embeddings; the latter is the strategy of most pre-trained word embeddings. This embedding is used to show that regional word embeddings perform differently for regionalized tasks.\footnote{These 27 embeddings are available in~\url{https://ingeotec.github.io/regional-spanish-models/}}


\begin{figure}[!ht]
    \centering
    \includegraphics[width=0.6\textwidth]{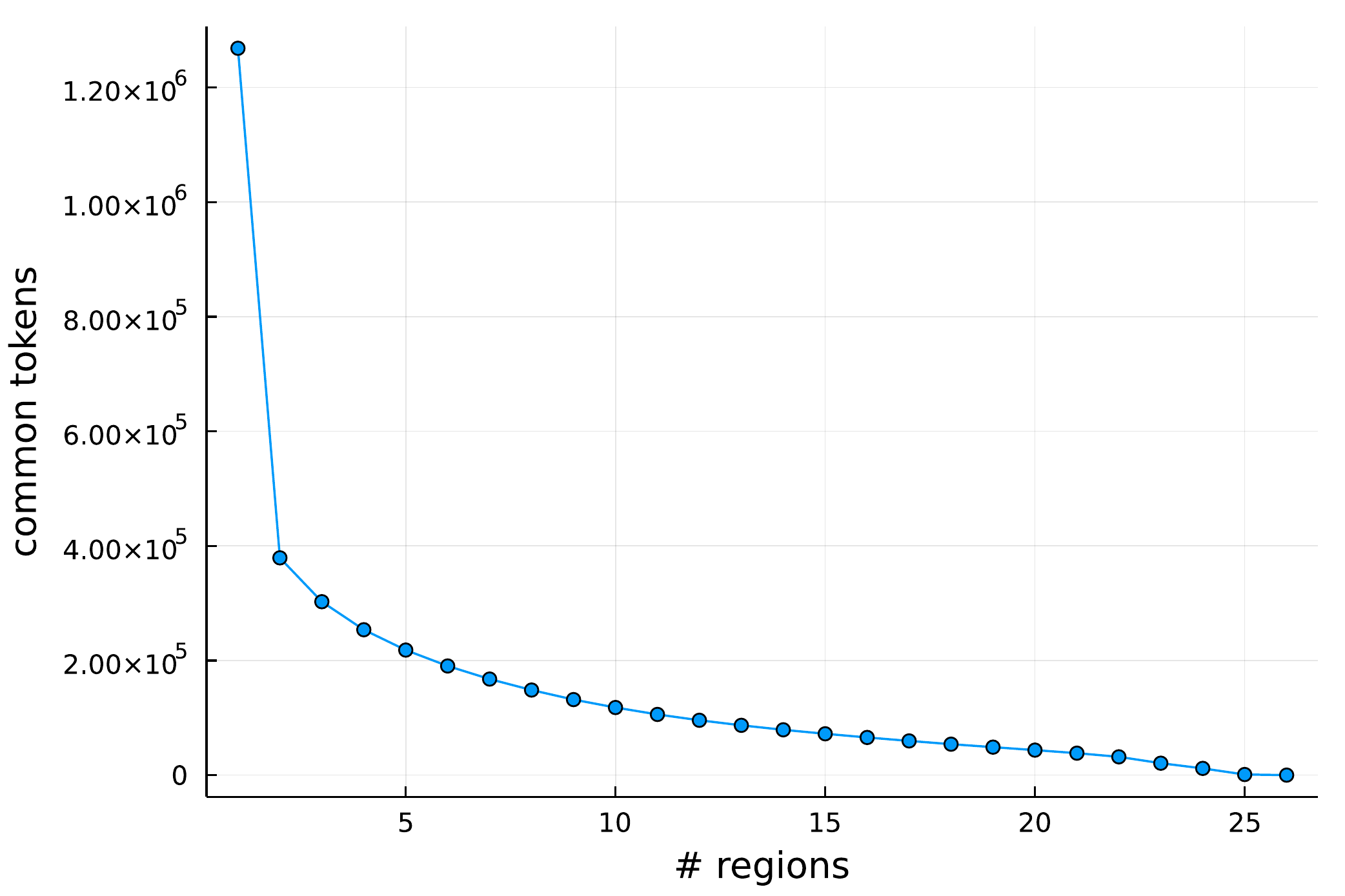}
    \caption{Number of common tokens shared by different countries or regions.}
    \label{fig/common-tokens}
\end{figure}

\subsection{Word-embedding similarity}
Our semantic analysis requires an affinity matrix, as the lexical one given in the previous section. Therefore, we need a representation and a similarity measure to compare word embeddings.
Please note that regular word embeddings produced with neural networks will generate vectors that cannot be mixed. Please recall that in the first stage of learning each neural network, its parameters are randomly initialized. An optimizing algorithm is then used to minimize a loss function on the dataset, adjusting parameters and iterating until some objective is achieved. These two procedures, random initialization and optimizing for different datasets, make that two neural network models produce no proximal vectors for the same word, i.e., under the cosine distance. Despite vectors having identical numerical structures, e.g., 300 dimensions, and components showing similar distributions, we cannot evaluate distances between points predicted in different models. 

We propose using an intermediate representation and a distance function that captures similarities between these embeddings to measure the similarity between different countries. The core idea is to represent each embedding with a flattened version of the $k$ nearest neighbor graph under a reduced set of tokens, i.e., tokens appearing in most word embeddings. Therefore, the similarity becomes linked to the neighborhood of each word (semantically similar words). The procedure to create this representation is the following:

\begin{itemize}
    \item Select a common set of tokens; each token appears in at least five countries. This filtering reduces the vocabulary from more than a million tokens to nearly 200 thousand tokens ($vocsize$). This selection corresponds to an inflection point in the tokens curve, (see Fig.~\ref{fig/common-tokens}). The core idea is to reduce the final representation dimensionality and increase the similarity between related words.
    \item Our representation requires constructing a $k$ nearest neighbor graph for each country. We use dense vectors of the word embeddings closed to the common tokens set, and we select $k=33$ after probing several choices. This $k$ value captures several similar terms and remains specific enough to let out different tokens. We used the cosine distance on the dense vectors.
    \item Finally, each country is represented as a high-dimensional vector that uses $k$ entries per vocabulary word, one per neighbor in the common token set. Each token is then represented by its $k$ nearest neighbors and weighted inversely to its distance.\footnote{We use the weighting form $0.5 + \frac{1}{1 + d(u, v)}$ for embedding vector $u$ and its neighbor's vector $v$, the lower the distance, the higher the weight.} Note that each word embedding is represented with a very sparse high-dimensional vector, i.e., $vocsize^2$ possible entries, and more than 3.8 million non-zero components.
\end{itemize}

\begin{figure}
    \centering
    \begin{subfigure}[b]{0.49\textwidth}
    \includegraphics[width=\textwidth]{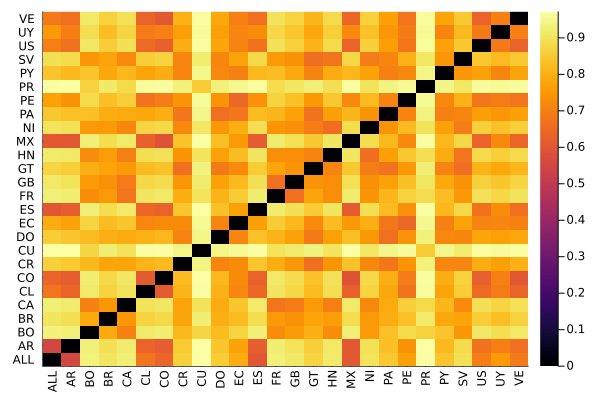}
    \caption{Affinity matrix of our semantic representations.}
    \label{fig/common-words-semantic-affinity-matrix}
\end{subfigure}~\begin{subfigure}[b]{0.48\textwidth}
    \fcolorbox{gray}{white}{\includegraphics[trim=0.5cm 0.25cm 0.5cm 0.2cm, clip, width=\textwidth]{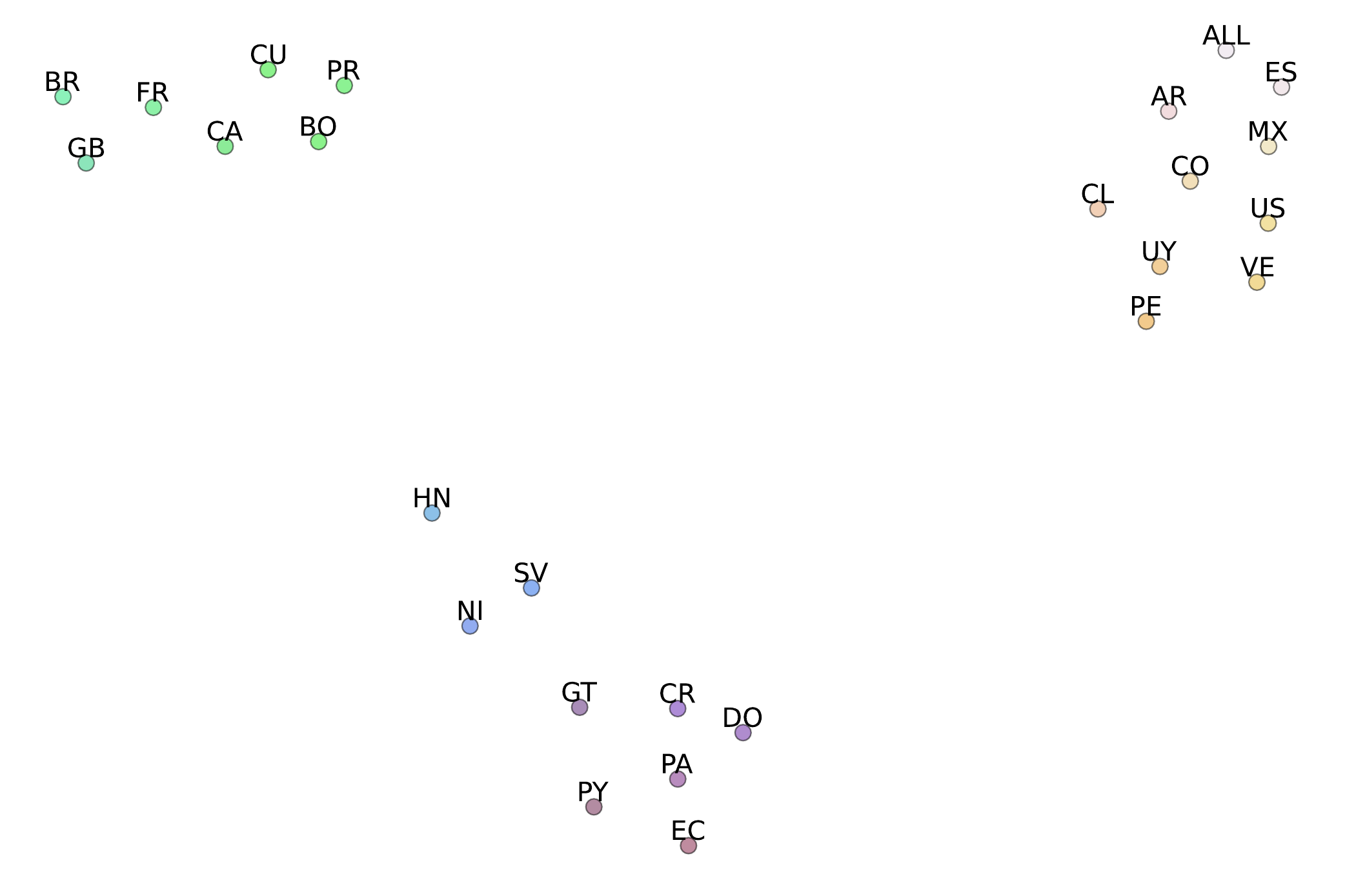}}
    \caption{Two dimensional UMAP projection of semantic representations.}
    \label{fig/semantic-umap}
    \end{subfigure}

\caption{Semantic similarities of our Spanish regional word embeddings. Countries are specified in their two letter ISO code. On the left, an affinity matrix where darker cells indicate higher similarities (small distances). On the right a two dimensional UMAP projection, near points indicate similarity.}
\end{figure}

The set of Spanish embeddings is compared with the cosine distance on the sparse vectors.
We computed the affinity matrix shown in Figure~\ref{fig/common-words-semantic-affinity-matrix} using the procedure described above. As in the previous affinity matrix, darker colors represent a high similarity between the regionalized embeddings and the contrary with lighter colors.

\begin{figure}[!ht]
\centering
    \includegraphics[width=\textwidth]{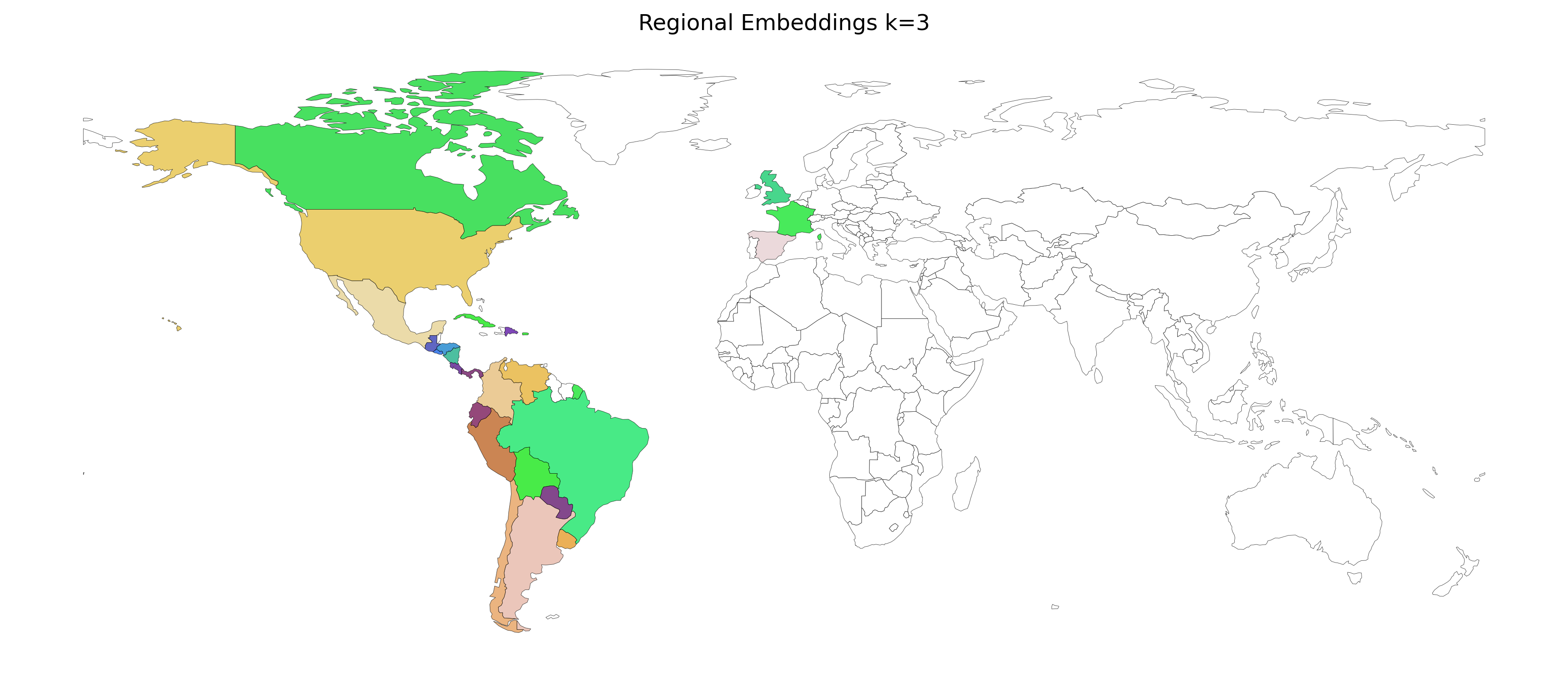}
    \caption{Geographic visualization of regional embeddings. The 3D UMAP projection is encoded as RGB.}
    \label{fig/geo-umap-semantic-colors}
\end{figure}

Figure~\ref{fig/semantic-umap} shows a two-dimensional UMAP projection of our semantic representation affinity matrix. The projection uses a $3$ nearest neighbors graph as input; please recall that few neighbors capture local structures.
The colors are computed by 3D reduction by applying the UMAP dimensional reduction to the same input; the resulting components create an RGB color set using a simple translation and scale procedure to compose values between $0$ and $1$. Both distances and colors describe a few well-defined groups. Please note that our {ALL} model is quite different from most points and similar to those regions with extensive collections. This effect is available in word embedding models constructed on non-regional corpora; they learn semantic traits of most represented regions. On the contrary, it is necessary to mention that countries with few messages (e.g., CU, PR, and BO) could need more data to support the learning procedure; nonetheless, we decided to maintain them in the projection to learn about their similarity, yet under this advisor. On the other hand, we remove GQ since their small vocabulary produces numerical errors while computing its corresponding $k$ nearest neighbor graph and UMAP projection.

Figure~\ref{fig/geo-umap-semantic-colors} shows the colormap computed from the previous dimensional reduction to colorize a world map obtaining a kind of map of semantic similarity of the Spanish language under our construction characteristics. We can observe how green colors group non-Spanish speaking countries (CA, FR, UK, and BR), except for PR, CU, and BO, in Fig.~\ref{fig/semantic-umap}. Note that they correspond to our corpus with fewer messages and that Fig.~\ref{fig:prop_tweets_in_country} also indicates a high number of foreign messages. It is necessary to take these results with reservation since can be issues related to their sizes, as explained in Section~\ref{sec:lexical-analysis} for the same set of collections.

Another large cluster is found with countries of all of Latin America (bottom of Fig.~\ref{fig/semantic-umap}). Here we see at least two subclusters with meaningful geographical meaning HN, NI, SV, and GT. The other groups include DO, CR, and PA. Note that EC and PY are also included here. Finally, we found a cluster containing countries around the world. Note that the ALL word embedding is also placed in this cluster. This cluster seems to be composed of countries with larger collections and other countries that are related to them. We found the AR, ES, CL, CO, UY, PE, US, VE, and MX here. Interestingly, we see the US here and not in the green cluster that agglutinates countries not having the Spanish language as an official or \textit{de facto} language.

The semantic similarities between word embeddings can be of interest and can be the object of further research, but their practical usage is also of interest. For instance, it is possible to know what countries can be mixed or interchanged without affecting the regional semantics significantly. A proper topic analysis could help clarify some of these clusters, but it is beyond the scope of this manuscript. 

\subsection{A regional task example: predicting emojis with Emoji-15}
\label{sec/emoji-15}
Regional information can be used to improve understanding of formal and informal messages, using typical terms and expressions in some regions but not necessarily used in others. Our regional models can help improve some NLP tasks having these characteristics.
We introduce the Emoji-15 classification task, a simple multiclass classification problem that predicts the emoji for given messages among 15 possible ones. This task involves identifying emotions and sentiments without a particular topic.

Its creation methodology is simple. We selected 15 popular emojis (see Section~\ref{sec:lexical-analysis}); we do not select the top 15 emojis per region, but a subset that gives some diversity in emotions. Also, we explicitly avoided the most popular emoji and skin tones from our selection. The selected emojis are listed in the first column of Table~\ref{tab:dist-emoji-15}. 
We selected the datasets for training and test sets from 2020's January and February; therefore, the corpus resources, training, and test sets are disjoint. We ensured that tweets contain at most one of these emojis (even when they can have other emojis). Messages were also selected to be geotagged to one of our objective Spanish-speaking countries. 
We followed the same filtering procedure and preprocessing as made for the word embedding; note that we also masked emoji's occurrences. That emoji was used as a label for the classification task.

We obtained a number of examples that were divided into a 50-50 holdout (proportion of label messages remain similar in train and test set); see Table~\ref{tab:dist-emoji-15} for more details. We removed four countries (BO, CU, GQ, and PR) from this task due to the low number of retrieved messages. For instance, we kept the statistics of Cuba in the table to show the lower limit cutting. The idea is to solve all-region benchmarks with all-region models and quantify their performance and the pertinence of local models on local tasks.

\begin{table}[ht]
\includegraphics[trim=2.3cm 9.7cm 2.3cm 1cm, clip, width=1\textwidth]{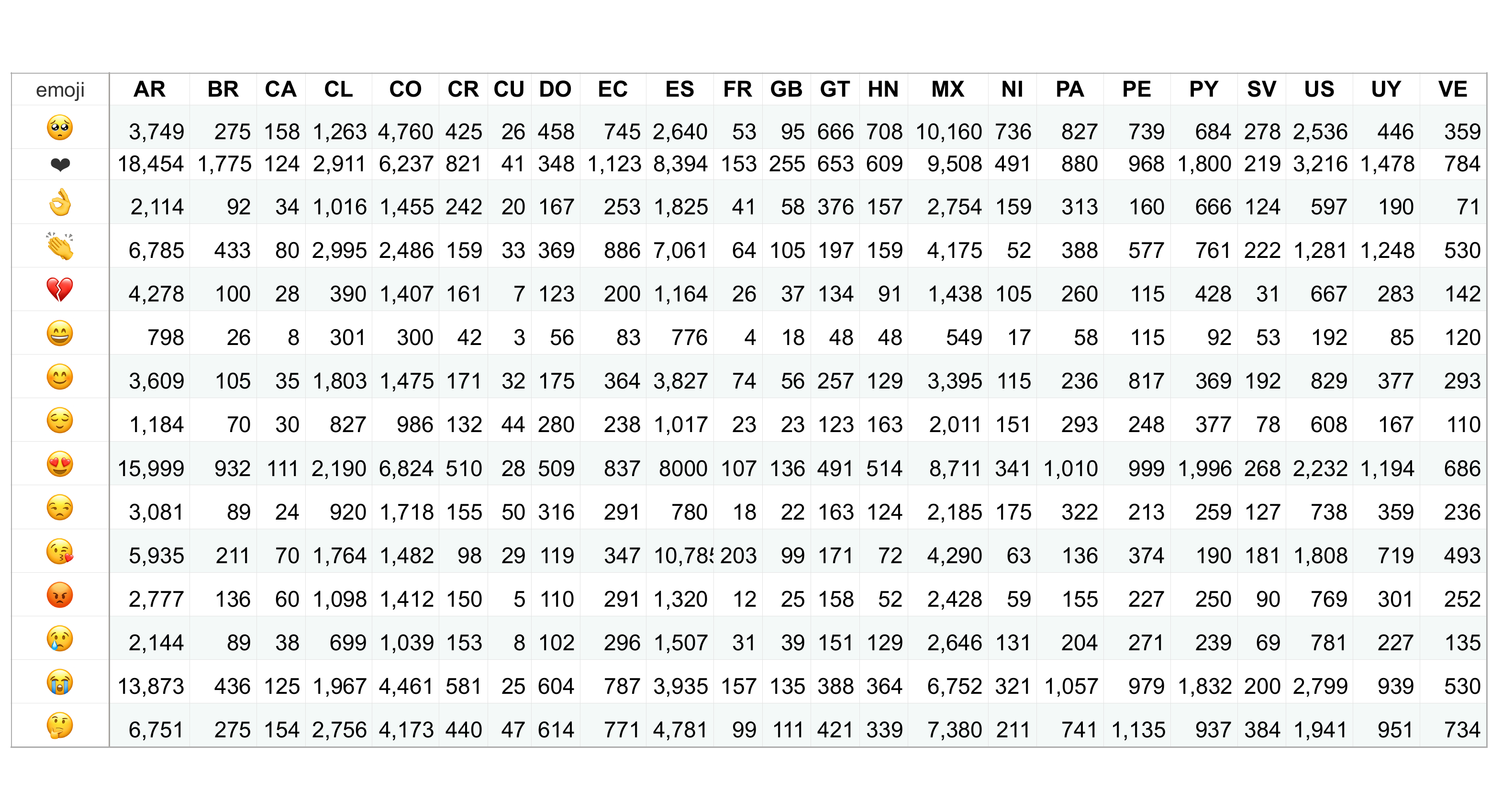}
    \caption{Train distribution of the emoji-15 datasets. Since it is a 50-50 hold-out partition, the test set follows a similar distribution. We removed countries with a low number of examples.}
    \label{tab:dist-emoji-15}
\end{table}

The train partition was used to create one model per country and one for the entire set of messages (called ALL). 
Table~\ref{tab/performance-emoji15} shows the accuracy performance scores of all models vs. all test databases. We can observe that some regions are more challenging to predict than others, e.g., CA achieves a maximum score of 0.35 while AR achieves 0.49. One can observe that most accuracy scores are low but far from a uniform distribution (15 classes).
The table shows the local model's position in each country benchmark (\textit{local rank} column). The best-performing model for a benchmark will rank as 1, the second-best as 2, and similarly for the rest. 
Note that small local rank values indicate that the local model (for that region) is efficient for its corresponding benchmark. The best five models for each country benchmark are also listed; we can observe how many geographically near regions perform well in their geographic neighborhoods. The average rank of the local model is 8.09 while the median is 6.5; these values support the idea that local models are useful on tasks where regional information can be used.
Also, one can observe that not always more data (ALL model) is the best, in this case, it could be said that the geographical aspect is more relevant. 


\begin{table}
    \centering
\begin{tabular}{crlr ccccc}
\toprule
\bf country & \bf min & \bf max & \bf local & \multicolumn{5}{c}{\bf top-5} \\ 
\bf code   & \bf acc. & \bf acc. & \bf rank & 1 & 2 & 3 & 4 & 5\\
\midrule
AR & 0.478 & 0.490 & 3 & UY & PY & AR & PE & CO\\
BR & 0.461 & 0.488 & 1 & BR & {ALL} & DO & PY & CR\\
CA & 0.293 & 0.353 & 18 & CL & {ALL} & CO & MX & US\\
CL & 0.426 & 0.449 & 1 & CL & US & MX & AR & ES\\
CO & 0.425 & 0.437 & 2 & US & CO & VE & EC & GT\\
CR & 0.369 & 0.388 & 9 & US & VE & {ALL} & MX & CO\\
DO & 0.338 & 0.381 & 13 & US & CO & VE & CL & {ALL}\\
EC & 0.380 & 0.414 & 9 & MX & US & CL & {ALL} & CO\\
ES & 0.475 & 0.486 & 1 & ES & AR & MX & US & VE\\
FR & 0.419 & 0.442 & 4 & {ALL} & GT & EC & FR & PA\\
GB & 0.347 & 0.376 & 23 & {ALL} & AR & ES & VE & MX\\
GT & 0.349 & 0.388 & 13 & MX & US & {ALL} & CO & ES\\
HN & 0.335 & 0.367 & 18 & PE & EC & BR & CR & UY\\
MX & 0.423 & 0.434 & 1 & MX & GT & CR & US & CO\\
NI & 0.337 & 0.372 & 18 & VE & CO & CL & MX & US\\
PA & 0.366 & 0.393 & 10 & US & CL & VE & CO & PE\\
PE & 0.380 & 0.420 & 9 & MX & {ALL} & US & AR & CO\\
PY & 0.424 & 0.442 & 1 & PY & US & BR & PE & UY\\
SV & 0.323 & 0.395 & 18 & US & CO & MX & CL & VE\\
US & 0.404 & 0.424 & 1 & US & MX & CO & ES & CL\\
UY & 0.435 & 0.457 & 1 & UY & US & CO & CL & VE\\
VE & 0.385 & 0.434 & 4 & MX & CO & ES & VE & US\\
\bottomrule
\end{tabular}
\caption{Performance statistics of all benchmarks (countries). Top-5 models are also listed and the rank position of the local model on solving the current benchmark.}
\label{tab/performance-emoji15}
\end{table}

\begin{table}[!h]
    \centering
\begin{tabular}{crr}
\toprule
\bf model & \bf voc. & \bf avg.\\
          & \bf size & \bf rank\\
\midrule
US & 292,465 & 4.23\\
CO & 324,635 & 6.05\\
MX & 438,136 & 6.27\\
CL & 282,737 & 6.91\\
VE & 271,924 & 7.00\\
ALL & 1,696,232 & 8.45\\
PE & 178,113 & 8.64\\
UY & 200,032 & 8.73\\
EC & 147,560 & 8.95\\
AR & 673,424 & 9.41\\
ES & 571,196 & 10.95\\
PY & 124,162 & 11.14\\
BR & 127,205 & 11.27\\
CR & 103,086 & 12.50\\
PA & 111,635 & 13.36\\
GT & 95,252 & 13.64\\
DO & 108,655 & 14.91\\
GB & 82,418 & 18.00\\
NI & 68,605 & 18.18\\
FR & 69,843 & 18.91\\
CA & 63,161 & 19.00\\
SV & 73,833 & 19.14\\
HN & 60,580 & 20.36\\
\bottomrule
\end{tabular}
\caption{Average rank of all regional models along all countries datasets. Models with low average ranks are better.}
\label{tab/avg-rank}
\end{table}

The average rank of a single model along all benchmarks indicates how well this model generalizes. Table~\ref{tab/avg-rank} shows the performance of all models, along with all benchmarks, as its average rank. We can observe that some country models are outstanding, like the US model. In this sense, it is remarkable that  models like the US or CO (both using a vocabulary of $300$k tokens) perform better than huge ones. On the other hand, the ALL model is competitive; however, it is not the best (global 6th regarding average rank).
Please recall that we created the ALL model by merging the entire corpora into a single corpus, which is the typical construction; for instance, the ALL model contains close to $1.7$ million tokens in its vocabulary.\footnote{Note that our vocabulary in Section~{\ref{sec:lexical-analysis}} has more than $1.2$ million tokens, and here we mentioned a larger one; this is the vocabulary recognized by the fastText parser. However, both vocabularies were computed using the same corpus. It is similar for other word embeddings listed in Table~\ref{tab/avg-rank}.}

While these results apply to the regional task of predicting the most popular emojis, the evidence points out that local models are competitive options for solving tasks requiring local traits as emoji predictions (see Table~\ref{tab/performance-emoji15}). Even more, some regional models perform better than large ones, as shown in Table~\ref{tab/avg-rank}, which is remarkable.

\FloatBarrier
\section{Language Models}
\label{sec/language-models}
Language Models (LM) are more sophisticated than word embedding models since they go beyond word semantics to context semantics and text generation. In more detail, Word2Vec, FastText, and Glove generate fixed embeddings for each word independently if the word can take different meanings depending on the context. For example, the word \textit{orange} can be a fruit or a color, depending on the context. Language Modeling is the task of predicting the next word given some context so they perform well in distinguishing homonyms.

In that sense, BERT~\cite{devlin-etal-2019-bert} is an LM that has gained considerable attention lately. It is a model that uses a series of encoders to generate embeddings for each word depending on its context. BERT differs from alternatives like ELMo~\cite{peters-etal-2018-deep} because the same pre-trained model can be fine-tuned for different tasks. The pre-train on BERT uses the Masked Language Model (MLM) task where each input sentence contains a \emph{mask token} on 15\% random words. Then, BERT was trained on a second task, the Next Sentence Prediction (NSP) task, where the input has two sentences, with a \emph{separation token} in between, and the task was to predict if the second sentence followed the first.

Our resources include regional pre-trained BERT-like models using the MLM task over tweets for the countries AR, CL, CO, MX, ES, UY, VE, and the US, i.e., larger ones. 
First, we applied the same preprocessing as detailed in Section~\ref{sec:lexical-analysis} to our corpora. The pre-training was the same as the original BERT, where 15\% of the tokens on each sentence were marked with a [MASK] token, and the model must predict them. We used the corresponding regional tweets from 2016 to 2019 to pre-train each model. All the models had a series of two encoders with four attention heads each and output 512-dimensional embedding vectors. This configuration corresponds with the small-size model following the official BERT implementation setup. We chose this setup based on the computational resources we had available. 
We name our model BILMA, for Bert In Latin America.
We used a learning rate of $10^{-5}$ with the Adam optimizer; the models for CL, UY, VE, and the US were trained for three epochs and AR, CO, MX, and ES for just one because of the size of their corpus. All the pre-trained models are available for download\footnote{\url{https://ingeotec.github.io/regional-spanish-models/}}.

\begin{figure}
\centering
\begin{subfigure}[t]{0.5\textwidth}
        {\includegraphics[clip=true, width=\textwidth]{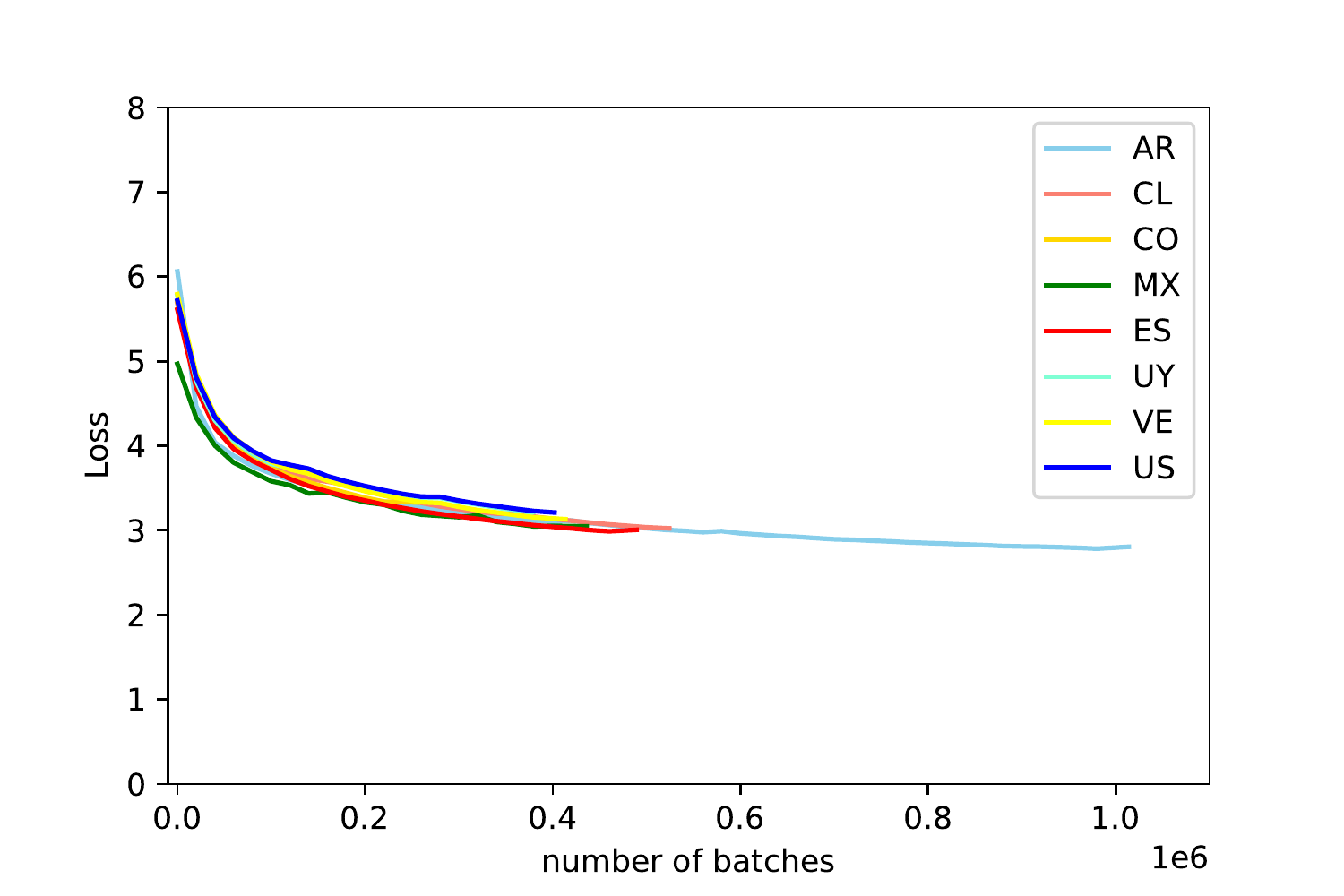}}
    \caption{Loss function values on training}
    \label{fig:bilma-loss}
\end{subfigure}~\begin{subfigure}[t]{0.5\textwidth}
    {\includegraphics[clip=true, width=\textwidth]{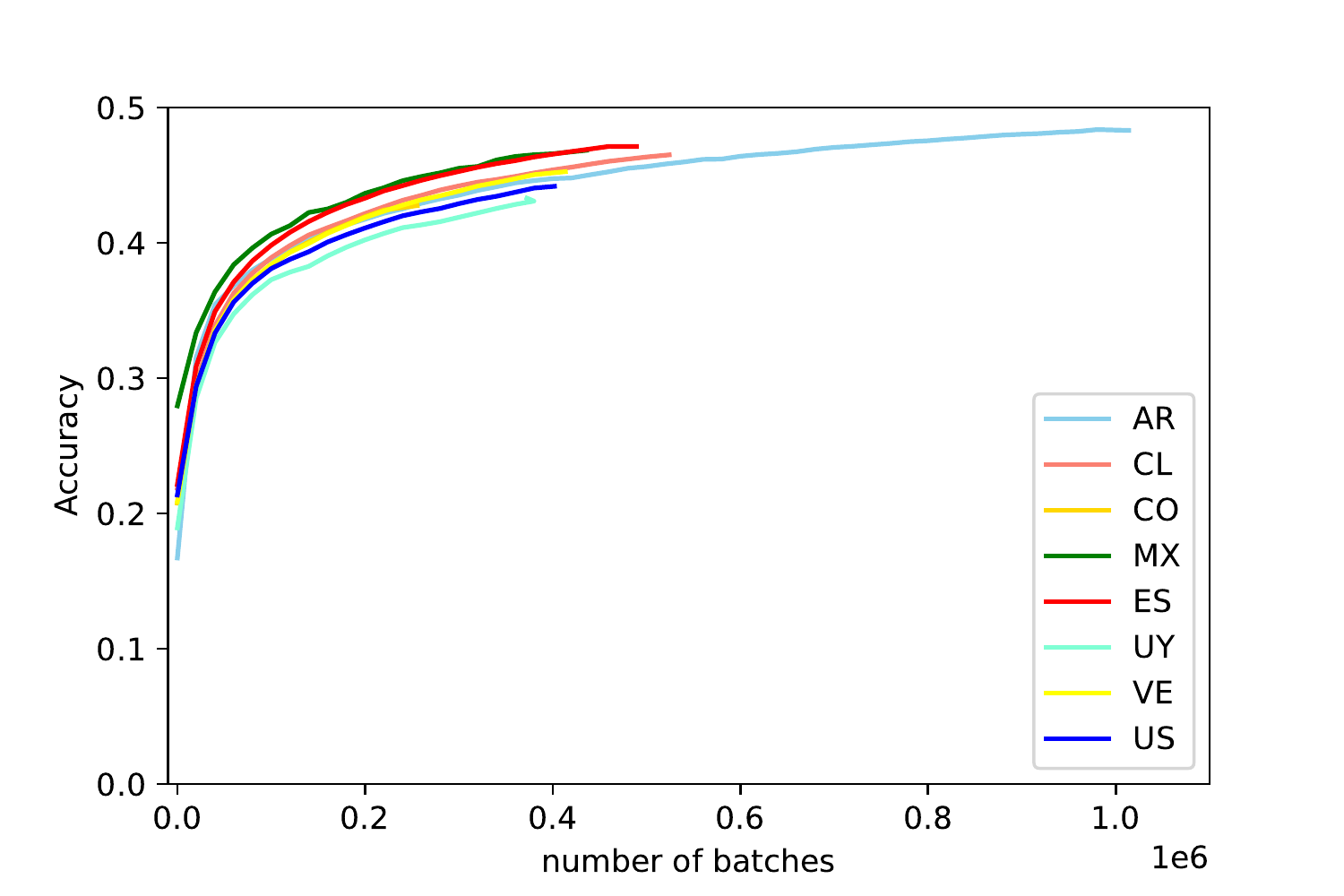}}
    \caption{Accuracy function values on training}
    \label{fig:bilma-acc}
\end{subfigure}
\caption{Loss and accuracy during training on the Masked Language Model task. The batch size is of 128 tweets.}
\label{fig:bilma-mlm-train}
\end{figure}

Figure~\ref{fig:bilma-mlm-train} shows the loss and accuracy of the MLM task during the training. We can see that the BILMA model for AR was trained on double the number of batches; that was because Argentina has double the corpus size. The rest of the models were trained on a similar number of batches.

\begin{figure}
\centering
\begin{subfigure}[t]{0.5\textwidth}
        {\includegraphics[clip=true, width=\textwidth]{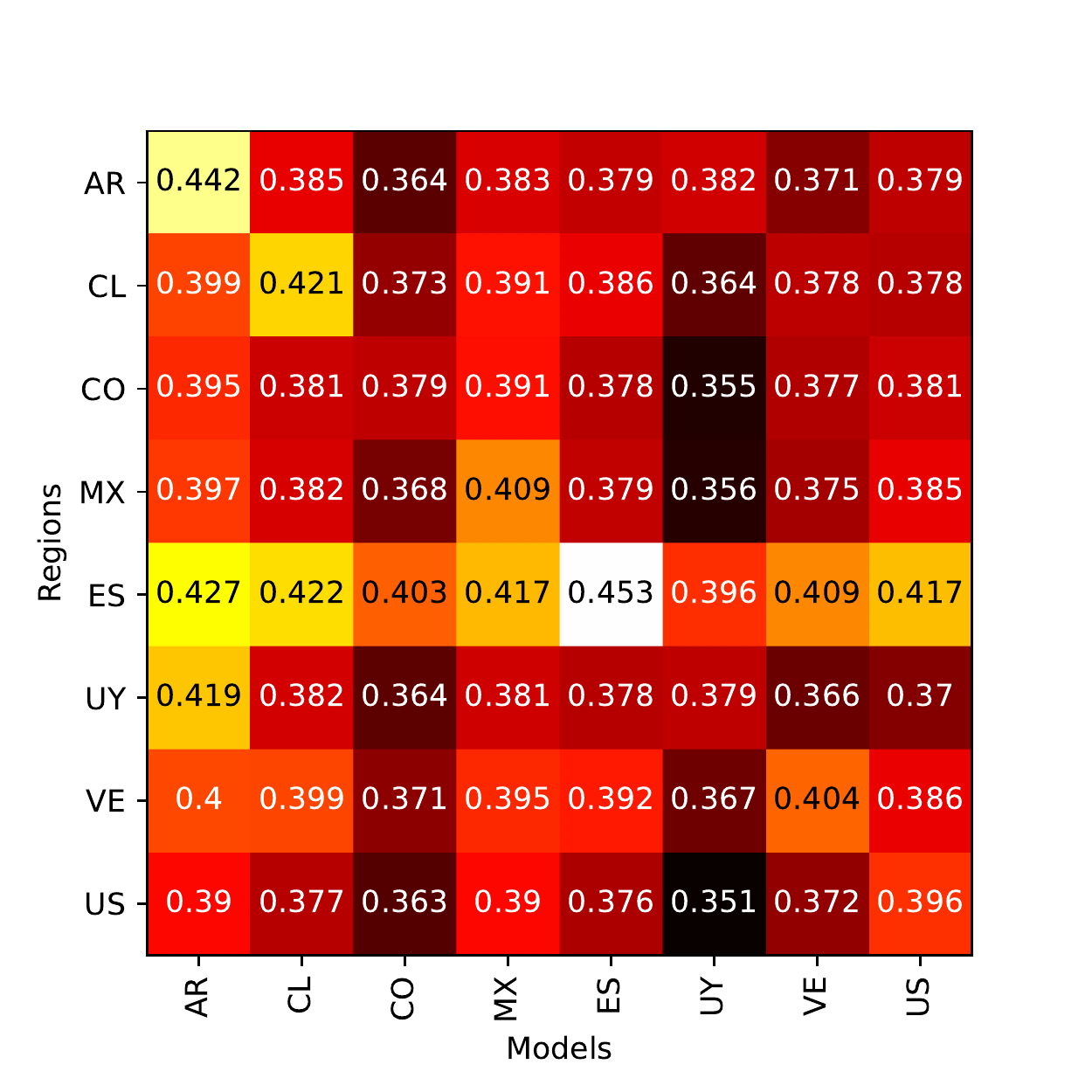}}
    \caption{Accuracy on the MLM task.}
    \label{fig:bilma-mlm}
\end{subfigure}~\begin{subfigure}[t]{0.5\textwidth}
    {\includegraphics[clip=true, width=\textwidth]{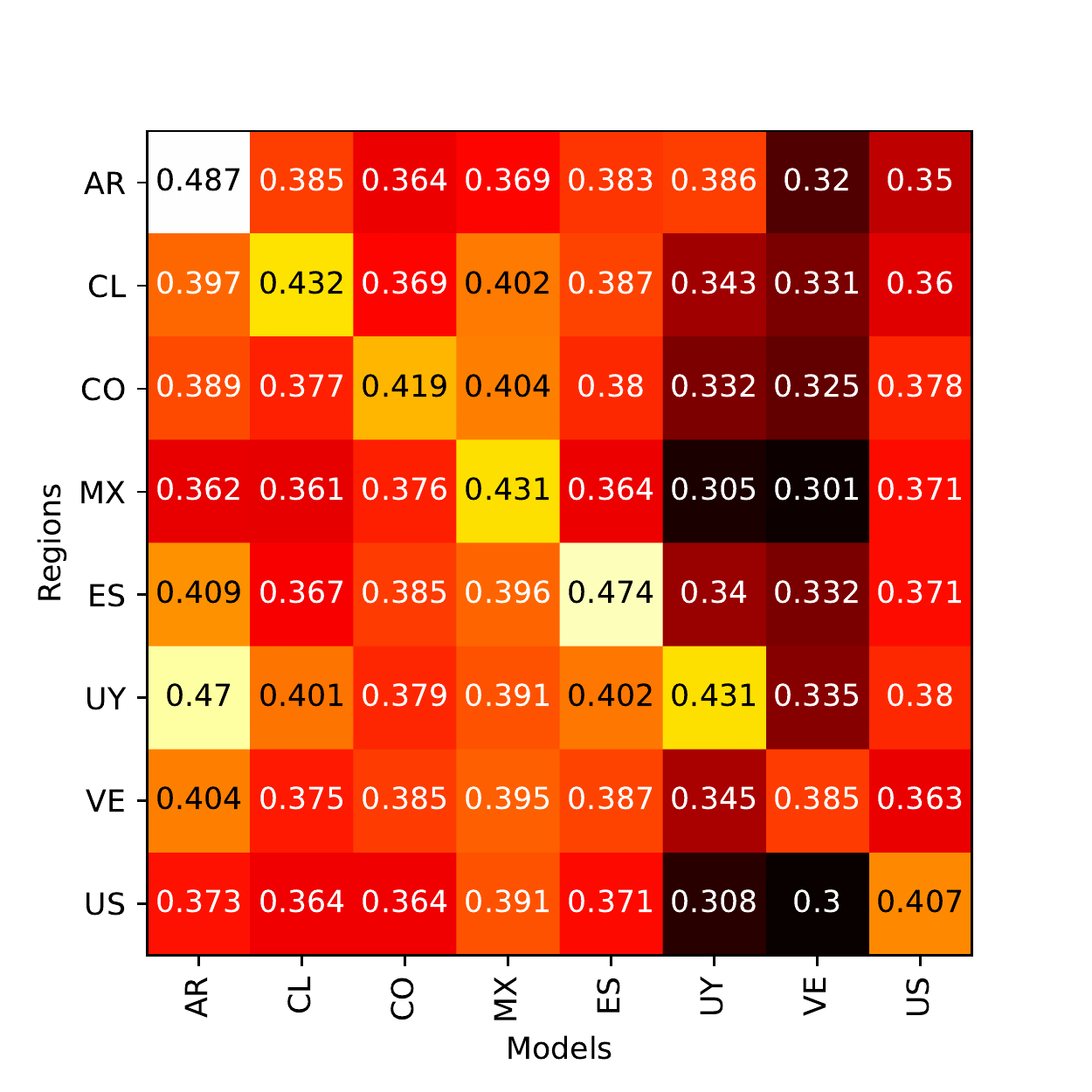}}
    \caption{Accuracy predicting the emoticon. }
    \label{fig:bilma-emo-pred}
\end{subfigure}
\caption{Comparison of the accuracy of the trained models on all the regions on MLM and emoticon prediction tasks.}
\label{fig:bilma-res}
\end{figure}

In Figure~\ref{fig:bilma-mlm}, we compare the models predicting the masked words on all the regions over the test set of tweets used in Section~\ref{sec/emoji-15}. The test was to predict the [MASK] tokens correctly. Some interesting points to highlight are the following. First, the Argentina model got very high scores on all the regions, even above their corresponding models for UY and CO. This might be because this model was trained for like double the data. Second, some models got better results in the ES region than theirs, like CL, CO, UY, VE, and the US. The US region got the worst outcomes for AR, CL, CO, ES, and UY models. Finally, CO and UY were the models with lower accuracy.

\subsection{BILMA's performance on the Emoji-15 regional task}
We applied our BILMA models to our Emoji-15 task (see Section~\ref{sec/emoji-15}). For this matter, we fine-tuned the pre-trained language models to predict the emoticon by adding two linear layers to the first token of each sentence (the start-of-sentence token), so the output of the fine-tuned models was a probability distribution of the assigned emoticon, independent of its position. We split the tweets into 90\% train and 10\% validation and trained until the accuracy stalled. After that, we evaluated the test set; the results are presented in Figure~\ref{fig:bilma-emo-pred}. We can conclude that all the models got better results in their corresponding regions from the results. The AR, MX, and ES models got good results over all the regions; meanwhile, UY and VE got low scores. 
The prediction scores are similar to those found in Table~\ref{tab/performance-emoji15}; however, our regional FastText models are slightly better than our fine-tuned BILMA models. Nonetheless, BILMA models learn how people write in different regions, as it is exemplified in the rest of this section.

\subsection{Generating text with BILMA regional language models.}
As a qualitative and exemplification exercise, we present how each region model predicts the masked word for the same example phrase. In Table~\ref{tab/bilma-predictions} we show the predictions for the masked token on a set of selected sentences. The color intensity indicates the confidence of the model to predict the word. The first two examples are \textit{el/la [MASK] subio de precio} (the [MASK] raised in price),\footnote{the article \textit{el} indicates the masked word should be singular and masculine, and for \textit{la} it should be singular feminine} here we can see differences in how each region name their public transportation, in AR they use \textit{bondi, colectivo}, in CL \textit{metro, micro, bus}, in MX \textit{uber} and ES \textit{metro, bus}; we can also see the differences in how they called the cellphone service, in CO and MX is \textit{celular} and in ES is \textit{movil}.
The third example is \textit{me gusta tomar [MASK] en la mañana} (I like to drink [MASK] in the morning)\footnote{\textit{tomar} could mean to take or to drink}, here we can note that in AR and UY people prefer to drink \textit{mates} meanwhile in MX, ES, VE, and the US drink coffee.
The fourth phrase is \textit{vamos a comer [MASK]} (let us eat [MASK]) where we can see the differences in the cuisine of the countries with dishes like \textit{asado, pizza, ñoquis, empanadas, milanesas, sushi, tacos, oreja, hamburguesa, arepa, torta}.
The last sentence is \textit{estoy en la ciudad de [MASK]} (I am in [MASK] city). The results include a list of some of the larger cities in each region. This exercise is a proof of concept to show that the models of different regions can predict very different words, i.e., regional information. Note that the diversity of the predictions include dialect differences (\textit{celular} vs \textit{movil}) but also topical (\textit{tacos} vs \textit{asado}) and will depend on the input sentence.

\begin{table}
\centering
\resizebox{4in}{!}{\begin{tabular}{|c|c|c|c|c|c|c|c|}
\hline
\multicolumn{8}{|c|}{\textbf{el [MASK] subio de precio}} \\ \hline
 AR & CL & CO & MX & ES & UY & VE & US \\ \hline
 \cellcolor{red!25}dolar & \cellcolor{red!15}chofer & \cellcolor{red!19}que & \cellcolor{red!15}cel & \cellcolor{red!25}que & \cellcolor{red!24}video & \cellcolor{red!15}internet & \cellcolor{red!57}que\\
\hline
 \cellcolor{red!19}bondi & \cellcolor{red!15}metro & \cellcolor{red!18}tiempo & \cellcolor{red!14}video & \cellcolor{red!23}movil & \cellcolor{red!12}que & \cellcolor{red!13}video & \cellcolor{red!6}video\\
\hline
 \cellcolor{red!13}que & \cellcolor{red!14}0 & \cellcolor{red!12}0 & \cellcolor{red!12}uber & \cellcolor{red!8}tiempo & \cellcolor{red!12}profe & \cellcolor{red!12}0 & \cellcolor{red!6}juego\\
\hline
 \cellcolor{red!6}0 & \cellcolor{red!13}que & \cellcolor{red!9}agua & \cellcolor{red!10}numero & \cellcolor{red!8}bus & \cellcolor{red!11}0 & \cellcolor{red!12}precio & \cellcolor{red!4}q\\
\hline
 \cellcolor{red!6}auto & \cellcolor{red!8}bus & \cellcolor{red!7}se & \cellcolor{red!9}celular & \cellcolor{red!6}dia & \cellcolor{red!10}horoscopo & \cellcolor{red!11}telefono & \cellcolor{red!4}0\\
\hline
 \cellcolor{red!6}video & \cellcolor{red!7}internet & \cellcolor{red!7}ano & \cellcolor{red!9}tuit & \cellcolor{red!6}metro & \cellcolor{red!7}tema & \cellcolor{red!8}que & \cellcolor{red!4}pueblo\\
\hline
 \cellcolor{red!6}autoestima & \cellcolor{red!6}papa & \cellcolor{red!6}dia & \cellcolor{red!8}que & \cellcolor{red!6}coche & \cellcolor{red!6}nombre & \cellcolor{red!7}dolar & \cellcolor{red!3}arbitro\\
\hline
 \cellcolor{red!5}colectivo & \cellcolor{red!6}mio & \cellcolor{red!6}man & \cellcolor{red!7}telefono & \cellcolor{red!5}agua & \cellcolor{red!5}pibe & \cellcolor{red!6}queso & \cellcolor{red!3}tipo\\
\hline
 \cellcolor{red!5}tiempo & \cellcolor{red!6}precio & \cellcolor{red!5}mundo & \cellcolor{red!6}dinero & \cellcolor{red!4}cafe & \cellcolor{red!5}grupo & \cellcolor{red!6}pan & \cellcolor{red!3}no\\
\hline
 \cellcolor{red!4}tren & \cellcolor{red!6}profe & \cellcolor{red!5}celular & \cellcolor{red!6}tiempo & \cellcolor{red!4}pelo & \cellcolor{red!4}amor & \cellcolor{red!5}tlf & \cellcolor{red!3}mundo\\
\hline
\multicolumn{8}{|c|}{\textbf{la [MASK] subio de precio}} \\ \hline
 AR & CL & CO & MX & ES & UY & VE & US \\ \hline
 \cellcolor{red!36}foto & \cellcolor{red!29}wea & \cellcolor{red!37}gente & \cellcolor{red!23}foto & \cellcolor{red!37}gente & \cellcolor{red!54}foto & \cellcolor{red!14}gente & \cellcolor{red!66}gente\\
\hline
 \cellcolor{red!15}que & \cellcolor{red!20}gente & \cellcolor{red!24}vida & \cellcolor{red!12}vida & \cellcolor{red!11}que & \cellcolor{red!9}vida & \cellcolor{red!13}plata & \cellcolor{red!12}que\\
\hline
 \cellcolor{red!10}lluvia & \cellcolor{red!10}micro & \cellcolor{red!5}noche & \cellcolor{red!12}gasolina & \cellcolor{red!10}foto & \cellcolor{red!8}gente & \cellcolor{red!12}semana & \cellcolor{red!4}foto\\
\hline
 \cellcolor{red!7}luna & \cellcolor{red!9}foto & \cellcolor{red!5}que & \cellcolor{red!10}cancion & \cellcolor{red!7}vida & \cellcolor{red!7}historia & \cellcolor{red!11}caja & \cellcolor{red!3}prensa\\
\hline
 \cellcolor{red!6}musica & \cellcolor{red!7}mama & \cellcolor{red!5}ley & \cellcolor{red!8}gente & \cellcolor{red!7}noche & \cellcolor{red!5}profe & \cellcolor{red!10}foto & \cellcolor{red!2}policia\\
\hline
 \cellcolor{red!5}caja & \cellcolor{red!6}luna & \cellcolor{red!5}historia & \cellcolor{red!7}noche & \cellcolor{red!7}camara & \cellcolor{red!4}abuela & \cellcolor{red!9}carne & \cellcolor{red!2}tipa\\
\hline
 \cellcolor{red!4}gente & \cellcolor{red!4}mina & \cellcolor{red!4}semana & \cellcolor{red!7}ropa & \cellcolor{red!5}cara & \cellcolor{red!4}cancion & \cellcolor{red!7}vida & \cellcolor{red!2}camara\\
\hline
 \cellcolor{red!4}camara & \cellcolor{red!4}senora & \cellcolor{red!4}lluvia & \cellcolor{red!7}app & \cellcolor{red!4}bateria & \cellcolor{red!2}que & \cellcolor{red!7}cola & \cellcolor{red!1}ley\\
\hline
 \cellcolor{red!3}coca & \cellcolor{red!4}profe & \cellcolor{red!3}luna & \cellcolor{red!5}pagina & \cellcolor{red!4}musica & \cellcolor{red!2}vieja & \cellcolor{red!7}arepa & \cellcolor{red!1}cara\\
\hline
 \cellcolor{red!3}heladera & \cellcolor{red!3}vieja & \cellcolor{red!3}musica & \cellcolor{red!5}morra & \cellcolor{red!4}semana & \cellcolor{red!2}madre & \cellcolor{red!6}pasta & \cellcolor{red!1}0\\
\hline
\multicolumn{8}{|c|}{\textbf{me gusta tomar [MASK] en la manana}} \\ \hline
 AR & CL & CO & MX & ES & UY & VE & US \\ \hline
 \cellcolor{red!59}mates & \cellcolor{red!46}desayuno & \cellcolor{red!41}fotos & \cellcolor{red!61}cafe & \cellcolor{red!41}cafe & \cellcolor{red!33}mates & \cellcolor{red!62}cafe & \cellcolor{red!69}cafe\\
\hline
 \cellcolor{red!10}teres & \cellcolor{red!19}once & \cellcolor{red!10}cafe & \cellcolor{red!11}fotos & \cellcolor{red!24}algo & \cellcolor{red!25}mate & \cellcolor{red!20}fotos & \cellcolor{red!5}decisiones\\
\hline
 \cellcolor{red!6}cafe & \cellcolor{red!9}cafe & \cellcolor{red!8}cerveza & \cellcolor{red!6}decisiones & \cellcolor{red!12}nota & \cellcolor{red!15}algo & \cellcolor{red!2}ron & \cellcolor{red!5}ropa\\
\hline
 \cellcolor{red!5}helado & \cellcolor{red!6}agua & \cellcolor{red!8}agua & \cellcolor{red!5}agua & \cellcolor{red!5}decisiones & \cellcolor{red!4}agua & \cellcolor{red!2}clases & \cellcolor{red!4}clases\\
\hline
 \cellcolor{red!4}sol & \cellcolor{red!3}clases & \cellcolor{red!6}peliculas & \cellcolor{red!4}alcohol & \cellcolor{red!4}cerveza & \cellcolor{red!4}cafe & \cellcolor{red!2}decisiones & \cellcolor{red!2}manana\\
\hline
 \cellcolor{red!3}fernet & \cellcolor{red!3}sol & \cellcolor{red!6}decisiones & \cellcolor{red!3}cerveza & \cellcolor{red!2}cola & \cellcolor{red!4}sol & \cellcolor{red!2}hoy & \cellcolor{red!2}fotos\\
\hline
 \cellcolor{red!2}mate & \cellcolor{red!3}almuerzo & \cellcolor{red!5}hoy & \cellcolor{red!3}clases & \cellcolor{red!2}sol & \cellcolor{red!3}vino & \cellcolor{red!2}anis & \cellcolor{red!2}sol\\
\hline
 \cellcolor{red!2}birra & \cellcolor{red!2}cerveza & \cellcolor{red!4}ropa & \cellcolor{red!1}hoy & \cellcolor{red!2}todo & \cellcolor{red!3}alcohol & \cellcolor{red!1}cartas & \cellcolor{red!2}agua\\
\hline
 \cellcolor{red!2}algo & \cellcolor{red!2}hoy & \cellcolor{red!4}frio & \cellcolor{red!1}tequila & \cellcolor{red!2}uno & \cellcolor{red!3}helado & \cellcolor{red!1}agua & \cellcolor{red!2}hoy\\
\hline
 \cellcolor{red!1}birras & \cellcolor{red!2}helado & \cellcolor{red!4}musica & \cellcolor{red!1}comida & \cellcolor{red!1}alcohol & \cellcolor{red!3}cerveza & \cellcolor{red!1}0 & \cellcolor{red!2}comida\\
\hline
\multicolumn{8}{|c|}{\textbf{vamos a comer [MASK]}} \\ \hline
 AR & CL & CO & MX & ES & UY & VE & US \\ \hline
 \cellcolor{red!27}asado & \cellcolor{red!19}sushi & \cellcolor{red!19}mierda & \cellcolor{red!28}tacos & \cellcolor{red!27}mierda & \cellcolor{red!44}algo & \cellcolor{red!39}pizza & \cellcolor{red!39}\_\\
\hline
 \cellcolor{red!19}pizza & \cellcolor{red!19}todo & \cellcolor{red!18}jaja & \cellcolor{red!14}mucho & \cellcolor{red!16}ya & \cellcolor{red!9}pizza & \cellcolor{red!10}mierda & \cellcolor{red!18}hoy\\
\hline
 \cellcolor{red!9}algo & \cellcolor{red!11}hoy & \cellcolor{red!12}! & \cellcolor{red!11}jaja & \cellcolor{red!12}mas & \cellcolor{red!8}hoy & \cellcolor{red!9}manana & \cellcolor{red!7}conmigo\\
\hline
 \cellcolor{red!8}helado & \cellcolor{red!11}mierda & \cellcolor{red!12}rico & \cellcolor{red!11}pizza & \cellcolor{red!8}esto & \cellcolor{red!8}jaja & \cellcolor{red!8}rico & \cellcolor{red!7}pizza\\
\hline
 \cellcolor{red!8}noquis & \cellcolor{red!10}algo & \cellcolor{red!9}. & \cellcolor{red!9}0 & \cellcolor{red!7}mucho & \cellcolor{red!7}helado & \cellcolor{red!6}hamburguesa & \cellcolor{red!6}manana\\
\hline
 \cellcolor{red!7}pizzas & \cellcolor{red!8}pizza & \cellcolor{red!7}\_ & \cellcolor{red!8}manana & \cellcolor{red!6}jaja & \cellcolor{red!4}oreja & \cellcolor{red!6}hoy & \cellcolor{red!5}todo\\
\hline
 \cellcolor{red!7}hoy & \cellcolor{red!7}! & \cellcolor{red!5}helado & \cellcolor{red!4}hoy & \cellcolor{red!5}hoy & \cellcolor{red!4}todo & \cellcolor{red!5}arepa & \cellcolor{red!4}mierda\\
\hline
 \cellcolor{red!4}empanadas & \cellcolor{red!4}jaja & \cellcolor{red!5}pizza & \cellcolor{red!4}taquitos & \cellcolor{red!5}tio & \cellcolor{red!4}nada & \cellcolor{red!5}. & \cellcolor{red!4}bien\\
\hline
 \cellcolor{red!3}facturas & \cellcolor{red!4}ctm & \cellcolor{red!4}hoy & \cellcolor{red!4}emo & \cellcolor{red!5}usr & \cellcolor{red!3}uno & \cellcolor{red!4}todo & \cellcolor{red!3}playa\\
\hline
 \cellcolor{red!2}milanesas & \cellcolor{red!3}emo & \cellcolor{red!4}asi & \cellcolor{red!4}algo & \cellcolor{red!5}bien & \cellcolor{red!3}eso & \cellcolor{red!3}torta & \cellcolor{red!3}tacos\\
\hline
\multicolumn{8}{|c|}{\textbf{estoy en la ciudad de [MASK]}} \\ \hline
 AR & CL & CO & MX & ES & UY & VE & US \\ \hline
 \cellcolor{red!45}mierda & \cellcolor{red!31}santiago & \cellcolor{red!35}bogota & \cellcolor{red!87}mexico & \cellcolor{red!23}madrid & \cellcolor{red!40}mierda & \cellcolor{red!34}venezuela & \cellcolor{red!27}\_\\
\hline
 \cellcolor{red!15}argentina & \cellcolor{red!13}vina & \cellcolor{red!21}colombia & \cellcolor{red!3}monterrey & \cellcolor{red!13}espana & \cellcolor{red!30}casa & \cellcolor{red!13}caracas & \cellcolor{red!11}mexico\\
\hline
 \cellcolor{red!10}cordoba & \cellcolor{red!10}conce & \cellcolor{red!11}cali & \cellcolor{red!1}guadalajara & \cellcolor{red!9}verdad & \cellcolor{red!4}uruguay & \cellcolor{red!12}valencia & \cellcolor{red!10}dios\\
\hline
 \cellcolor{red!7}rosario & \cellcolor{red!9}chile & \cellcolor{red!9}medellin & \cellcolor{red!1}puebla & \cellcolor{red!9}sevilla & \cellcolor{red!3}hoy & \cellcolor{red!10}merida & \cellcolor{red!9}miami\\
\hline
 \cellcolor{red!3}mierdaa & \cellcolor{red!7}valparaiso & \cellcolor{red!8}barranquilla & \cellcolor{red!1}cancun & \cellcolor{red!9}barcelona & \cellcolor{red!3}historia & \cellcolor{red!8}barquisimeto & \cellcolor{red!8}hoy\\
\hline
 \cellcolor{red!3}tucuman & \cellcolor{red!7}stgo & \cellcolor{red!6}dios & \cellcolor{red!1}veracruz & \cellcolor{red!9}mierda & \cellcolor{red!3}nuevo & \cellcolor{red!5}maracaibo & \cellcolor{red!8}casa\\
\hline
 \cellcolor{red!3}hoy & \cellcolor{red!6}concepcion & \cellcolor{red!2}hoy & \cellcolor{red!0}mty & \cellcolor{red!7}hoy & \cellcolor{red!3}filosofia & \cellcolor{red!5}maracay & \cellcolor{red!7}vacaciones\\
\hline
 \cellcolor{red!3}capital & \cellcolor{red!5}valpo & \cellcolor{red!1}antioquia & \cellcolor{red!0}merida & \cellcolor{red!6}valencia & \cellcolor{red!3}ingles & \cellcolor{red!3}margarita & \cellcolor{red!6}pr\\
\hline
 \cellcolor{red!3}aca & \cellcolor{red!4}maipu & \cellcolor{red!1}cartagena & \cellcolor{red!0}cdmx & \cellcolor{red!5}futbol & \cellcolor{red!3}montevideo & \cellcolor{red!3}aragua & \cellcolor{red!5}disney\\
\hline
 \cellcolor{red!3}sol & \cellcolor{red!4}chillan & \cellcolor{red!1}paz & \cellcolor{red!0}todos & \cellcolor{red!5}examenes & \cellcolor{red!3}clase & \cellcolor{red!2}carabobo & \cellcolor{red!4}mi\\
\hline
\end{tabular}}
\caption{Predictions of the masked words over different regions. The color intensity indicates the probability of prediction.}
\label{tab/bilma-predictions}
\end{table}

\section{Conclusions}
\label{sec:conclusions}
This manuscript proposes a set of regionalized resources for the Spanish language using Twitter as the data source. We collected messages from Twitter's public streaming API from 2016 to 2019; messages must be tagged as being written in Spanish and geotagged to one of the 26 countries that use Spanish as one of their primary languages. 
The vocabulary of each corpus was extracted, characterized, and compared their similarity, defining a distance metric between them. We also produce visualizations and insightful information about lexical and semantical similarities of the Spanish language variations in Twitter messages. 

On the other hand, we created regional semantic models using FastText and produced some visualizations of the semantic similarities among regions. We also create regional language models called BILMA, based on the well-known BERT transformer architecture. We give empirical evidence of the usefulness of regional models in regionalized text classification tasks (Emoji-15 task) and how this more careful data segmentation can yield better performances than the typical more-data-is-better approach.

We provide access to our vocabularies, word embeddings, language models, and corpora sample through the project site (available in \url{https://ingeotec.github.io/regional-spanish-models/}). The necessary packages (BILMA) and all scripts used to generate our resources, open-sourced under the MIT license, are also reachable under the same site.

\subsection{Limitations and further research}
While the regional models seem promising tools for many tasks that require understanding regionalisms and idiosyncrasies, the use of multiple models can be cumbersome for real-world systems, not to mention the necessary computing units needed to handle many models. It is necessary to create models that can {\em shift} their region depending on a {\em regional context}. This approach requires further research.

Our Spanish corpora could be more balanced concerning countries. Some countries have too many elements, while others like GQ, CU, PR, and BO need to be bigger to have reliable semantic models (i.e., word embeddings and language models). More research and data collection are needed to improve resources in these regions.

Our region similarity comparisons are based on lexical and semantic properties of vocabularies computed and learned from Twitter messages. While it is not our goal, the presented projections could compare topics and other internal knowledge in the resources. Proper topic analysis is beyond the scope of this study and requires further research.
Similarly, it is possible to mine our resources, i.e., language models, semantic, lexical features, corpus, etc., to perform data-driven social sciences research, i.e., research about gender inequality or race perception. All these topics require more research.

The methodology and implementation of this manuscript are open to improvements. For instance, countries with relatively few examples require different strategies to be competitive. It is also essential to find ways to collect better data and discard the bad ones, such that results become more reliable for small collections. In the same sense, comparing vocabularies and embeddings created from datasets with such disparate sizes require robust normalization methods that we barely sketched. Our long-term goal is to update our resources using more and more data and novel language models as they appear in the literature.

\bmhead{Acknowledgments}
This work has been done through CONACYT (National Council of Science and Technology from Mexico) support with the Ciencia Básica grant with project ID A1-S-34811. Also, the authors acknowledge the support from CIMAT and ``Laboratorio de Supercómputo del Bajio'' through project 300832 from CONACyT. We also thank the reviewers' comments and suggestions that help us improve the manuscript's quality. The first author thanks CONACyT and CICESE (México) since this work was partially developed when the first author was in a fellow year visiting the latter institution.

\appendix
\section{BILMA language model usage}
\label{sec:apendice}

In order to use our BILMA models, we need to download one first, we will also need the vocabulary file.

To clone the repository, download the model and install dependencies, in a linux terminal just type the following commands:
\begin{lstlisting}[language=bash]
git clone https://github.com/msubrayada/bilma
cd bilma
bash download-emoji15-bilma.sh
python3 -m pip install tensorflow==2.4 
\end{lstlisting}
\noindent and now we have the python package and its dependencies, the model and the vocabulary file (shared to all BILMA models). In particular, this example downloads the MX model that was trained with one epoch on the MLM task and fine-tuned on the Emoji-15 task for 13 epochs.

We need to run a Python 3 console and load the BILMA model.

\begin{lstlisting}
from bilma import bilma_model
vocab_file = "vocab_file_All.txt"
model_file = "bilma_small_MX_epoch-1_classification_epochs-13.h5"
model = bilma_model.load(model_file)
tokenizer = bilma_model.tokenizer(vocab_file=vocab_file, max_length=280)
\end{lstlisting}

\noindent this BILMA model has two outputs, the first with shape \texttt{(bs, 280, 29025)} where \texttt{bs} is the batch size, 280 is the max length and 29025 is the size of the vocabulary. This output is used to predict the masked words. The second output has shape \texttt{(bs, 15)} which corresponds to the predicted emoji.

The next step is tokenizing some messages as follows:
\begin{lstlisting}
texts = [
 "Tenemos tres dias sin internet ni senal de celular en el pueblo.",
 "Incomunicados en el siglo XXI tampoco hay servicio de telefonia fija",
 "Vamos a comer unos tacos",
 "Los del banco no dejan de llamarme"
]
toks = tokenizer.tokenize(texts)
\end{lstlisting}
\noindent the prediction is made as follows:

\begin{lstlisting}
p = model.predict(toks)
\end{lstlisting}
\noindent finally, the predicted emojis can be displayed with:
\begin{lstlisting}
tokenizer.decode_emo(p[1])
\end{lstlisting}

\noindent this produces the output: \img{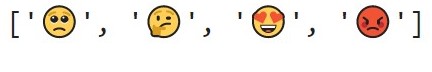}, each emoji corresponds to the most probable one for each message in \texttt{texts}.

\section{Cut off $N$}
\label{sec:cutoff}

This section presents a methodology that addresses the minimum token frequency to be kept in the analysis. 

The idea is to compute the confidence interval of a Bernoulli variable and select the minimum frequency $f$ (number of times the token appears in the corpus) that sets the interval in a feasible region. Let $p$ be the probability of seeing a particular token, assuming $\hat p$ is Gaussian distributed. The confidence interval is $\hat p \pm \alpha \textbf{se}(\hat p),$ where $\textbf{se}$ is the standard error of $\hat p$, and $\alpha$ is the percent point function with parameter $1 - \frac{c}{2},$ where $1 - c$ represents the confidence, e.g., $\alpha \approx 2$ gives approximately a 95\%
confidence interval.

The following equations show that under the assumption made, the frequency $f$ (number of times the token appears in the corpus) must be greater or equal to $\frac{N \alpha^2}{N + \alpha^2}$ that in the limit when $N$ tends to infinity corresponds to $\alpha^2$.

\begin{eqnarray}
    \hat p - \alpha \textbf{se}(\hat p) &\geq& 0 \\
    \hat p - \alpha \frac{\sqrt{\hat p (1 - \hat{p})}}{\sqrt{N}} &\geq& 0 \\
    \sqrt{N} \hat p - \alpha \sqrt{\hat p (1 - \hat{p})} &\geq& 0 \\
    \sqrt{N} \hat p &\geq&  \alpha \sqrt{\hat p (1 - \hat{p})} \\
    \sqrt{N} \hat p &\geq&   \sqrt{\alpha^2 \hat p (1 - \hat{p})} \\
    N \hat p^2 &\geq&   \alpha^2 \hat p (1 - \hat{p}) \\
    N \hat p^2 - \alpha^2 \hat p (1 - \hat{p}) &\geq& 0 \\ 
    \hat p (N \hat p - \alpha^2 (1 - \hat{p})) &\geq& 0 \\
    N \hat p - \alpha^2 (1 - \hat{p}) &\geq& 0 \\
    N \hat p - \alpha^2 + \alpha^2 \hat{p} &\geq& 0 \\
    \hat p (N + \alpha^2) &\geq& \alpha^2 \\
    \hat p &\geq& \frac{\alpha^2}{N + \alpha^2} \\
    f &\geq& \frac{N \alpha^2}{N + \alpha^2}
\end{eqnarray}

\end{document}